\documentclass[10pt,twocolumn,letterpaper]{article}

\usepackage[pagenumbers]{cvpr} % To force page numbers, e.g. for an arXiv version

\usepackage{cuted} %< for strip environment

\usepackage{currfile} %< for \currfilebase macro

\usepackage{caption} %< for "\captionof" commands (teaser)
\usepackage{array}     % ragged-right p{} columns in supplementary tables
\usepackage{placeins}  % \FloatBarrier in the supplementary material

\definecolor{cvprblue}{rgb}{0.21,0.49,0.74}
\usepackage[pagebackref,breaklinks,colorlinks,allcolors=cvprblue]{hyperref}

\title{Quantitative Motion Memory: Enabling Vision-Language Models \\ to Measure Motion Across Videos and Time}

\author{
Kyoungjin Jung$^{1}$ \quad
Heeyun Park$^{2}$ \quad
Jonghyuk Park$^{3}$ \quad
Dongwook Shin$^{1}$
}

\begin{document}
\maketitle

\begingroup
\renewcommand{\thefootnote}{}
\footnotetext{%
\footnotesize
\begin{tabular}{@{}r@{\hspace{0.7em}}l@{}}
\textsuperscript{1} &
Department of Mathematics, Ajou University \\
& \texttt{\{fned49,dws\}@ajou.ac.kr} \\
\textsuperscript{2} &
Independent Researcher, \texttt{gmldbs30113@gmail.com} \\
\textsuperscript{3} &
Department of AI, Big Data \& Management, Kookmin University \\
& \texttt{jonghyuk@kookmin.ac.kr}
\end{tabular}%
}
\endgroup

\begin{abstract}
Vision-language models (VLMs) describe what people do in a video, but they do not reliably measure how people move.
Asked which video shows the fastest hand movement, or when the hands rise highest in a video of several activities, a VLM judging RGB frames estimates the answer visually.
We propose Quantitative Motion Memory (QMM), a general training-free design in which a VLM answers from values measured on an explicit, time-indexed record.
The record is estimated once per video from ordinary monocular RGB, without motion capture or other sensors.
Operators apply measurement definitions shared by all records, so a value has the same meaning in every video and interval.
The VLM chooses what to measure and where, and views frames for visual judgments.
We instantiate QMM for human motion with estimated joint coordinates.
Against the datasets' 3D references and across seven VLMs, QMM raises accuracy from 2--40\% to 95--98\% in finding the interval where a quantity peaks in untrimmed videos of several activities, and from 7--49\% to 85--95\% on the quantitative steps of multi-step questions over videos from different datasets.
Without QMM, VLMs find at most 82\% of these intervals even with 55 times the input of QMM, and with QMM even the smallest evaluated VLMs outperform the strongest one without it on both tasks.
On independent human judgments, QMM agrees with 92--94\% of VidDiff annotations and with raters on 90--91\% of clip pairs from in-the-wild TED talks.
Ablations show that the record allows the VLM to locate intervals, the definitions make cross-video comparisons valid, and the operators make both hold across VLMs.
\end{abstract}

\section{Introduction}\label{sec:intro} 

\begin{figure*}[t!]
    \centering
    \includegraphics[width=\textwidth]{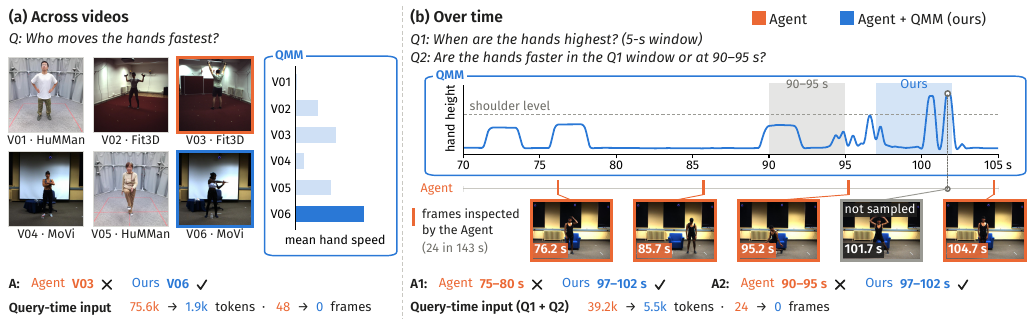}
    \caption{
        \textbf{Measuring from the record succeeds where judging from frames fails.}
        (a) Among six clips from three datasets, the VLM without QMM (Agent) selects V03 from eight frames per clip, whereas with QMM (Ours) it selects V06 from measured hand speeds.
        (b) In a 143-s video, the hands rise highest at 101.7\,s, between two of the Agent's 24 frames, and QMM retrieves the window around this peak from the record and reuses it for Q2.
        Both examples use GPT-6.1 Sol at low effort. Marks show agreement with the dataset-3D references, and the gray frame, which the Agent did not sample, is shown for reference.
}
    \label{fig:teaser}
\end{figure*}

Recent vision-language models (VLMs) have substantially improved video understanding~\cite{qwen35_2026,gemma4_2026,worldmm,videoarm}, yet many applications, such as fitness feedback, coaching, and gesture research~\cite{aifit2021,coachme2025,trujillo2019}, ask how people move rather than what they do.
Such quantitative motion analysis measures properties such as speed, height, or spacing over relevant intervals, compares them across people and videos, and answers multi-step questions whose later steps build on earlier answers.
Figure~\ref{fig:teaser} shows two examples, selecting the fastest hand movement among clips from different datasets and locating the window with the highest hands in a video of over two minutes.

Existing approaches address parts of this problem. VLM agents reason over sampled frames~\cite{videoagent_wang2024}, memory-augmented agents retrieve stored summaries or object memories~\cite{videoagent_fan2024,worldmm,videoemem2026}, and code-writing agents compute measurements from stored kinematics~\cite{amadeusgpt2023,biomechagent2026}.
Motion-specific methods describe single sequences numerically~\cite{smd2026}, answer categorical questions with programs over skeleton sequences~\cite{nspose2023,imore2025}, or verify instructions for a video pair with kinematic measurements~\cite{motionhalluc}.
Some of these approaches judge quantities visually, which is often unreliable~\cite{quantiphy,motionhalluc}. Others fix coarse descriptions before the question is known or measure within one sequence or video pair. These choices leave two limitations, one across videos and one over time.

First, quantities judged from frames are not reliably comparable across videos. Code written from scratch for each question is no more reliable, because it measures its own version of each quantity, such as hand speed without removing body motion (Sec.~\ref{sec:why_qmm}).
A VLM viewing frames estimates each quantity from a few images, so it is least accurate on close differences, changes its answers when asked again, and remains far less accurate as the collection grows (Sec.~\ref{sec:across_video_results}).

Second, temporal questions first require finding the relevant interval, which an agent viewing frames must do by scanning the video again for each question.
Sparse sampling misses brief events, as in Fig.~\ref{fig:teaser}(b). Searching a record remains more accurate than denser sampling, whole videos, the definitions given as text, or temporal grounding (Sec.~\ref{sec:temporal_results}).

\begin{figure*}[t!]
\centering
\includegraphics[width=\textwidth]{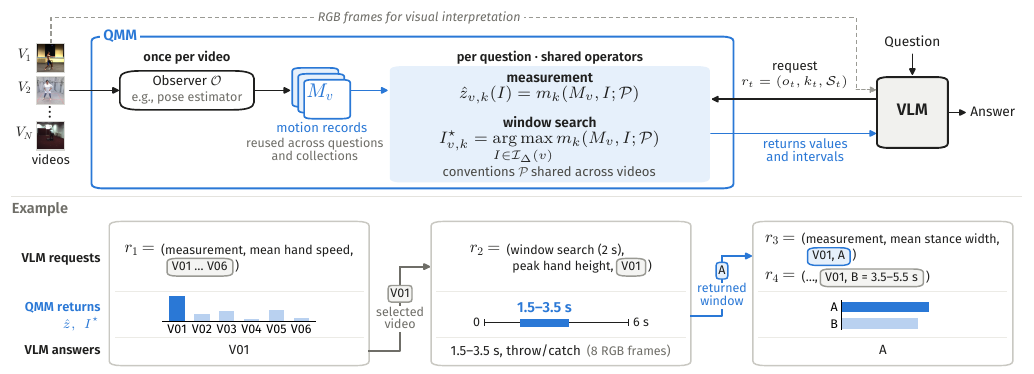}
\caption{
\textbf{Overview of Quantitative Motion Memory.}
An observer (here a pose estimator) estimates a motion record once per video, which is reused across questions.
For each question, the VLM requests measurements and window searches from operators that apply the shared measurement definitions, passes returned videos and windows to later requests, and inspects frames when needed.
Bottom, an evaluated multi-step question for GPT-6.1 Sol at low effort on a mixed-source collection, in which $r_3$ and $r_4$ measure stance width (Suppl.~Sec.~F).
}
\label{fig:framework}
\end{figure*}

We propose Quantitative Motion Memory (QMM), in which a VLM answers such quantitative questions from values measured on an explicit, time-indexed record estimated from a monocular RGB video. Measurement definitions shared across videos, namely the formulas and conventions of each quantity, give each value the same meaning in every video and interval. Perception builds the record once as a memory reused across questions, and the VLM decides what to compute. We instantiate QMM for human motion.
In this instantiation, QMM estimates a time-indexed record of joint coordinates from each monocular RGB video. Shared definitions, implemented once as operators, turn these coordinates into quantities such as body-relative hand speed that are defined identically in every video and can be computed for any interval. Searching the record by a quantity then locates intervals without viewing frames, and the VLM combines the results with the frames it inspects (Fig.~\ref{fig:framework}).
We further check the answers against raters who judge comparisons and windows without our formulas, including talks recorded in the wild, and against VidDiff annotations (Secs.~\ref{sec:across_video_results} and~\ref{sec:temporal_results}). Records from a second estimator test the dependence on the estimator (Sec.~\ref{sec:across_video_results}).

Our contributions are as follows.
\begin{itemize}\setlength{\itemsep}{0pt}\setlength{\parskip}{0pt}
\item \textbf{Quantitative Motion Memory.} QMM is a general training-free design in which an explicit, time-indexed record is estimated once per video and the VLM decides, for each question, what to measure on the record. Operators compute every value under measurement definitions shared across videos. Window search ranks the intervals of a video by a measured quantity, and the VLM views frames for visual judgments. For human motion, the record holds joint coordinates estimated from monocular RGB.
\item \textbf{What a VLM needs to measure motion.} Component ablations show what each part adds. The time-indexed record allows the VLM to find intervals that sampled frames miss. Without the shared definitions, even the strongest evaluated VLM writes code over the record that compares videos no more accurately than frames do. The operators give every VLM the same values, with less input than code written by the VLM. Records reduced in advance to whole-video values or peaks fall short, so the record must keep the temporal detail that later questions need.
\item \textbf{Evaluation across videos and time.} We evaluate seven VLMs, from 4B open models to commercial ones. QMM gives every VLM large gains on window search and on multi-step questions across datasets, and on window search and the quantitative steps of these questions the two smallest evaluated VLMs with QMM exceed the strongest VLM without it. QMM also locates intervals more accurately than a temporal-grounding model or the same VLMs given more frames, whole videos, or the definitions as text. On independent human judgments, its answers agree with VidDiff annotations and with raters on TED clip pairs recorded in the wild.
\end{itemize}
Our code, the evaluation questions with reference answers, and the anonymized rater answers will be made publicly available, together with motion records or scripts that rebuild them, as each dataset's license allows.

\section{Related Work}\label{sec:related_work}

\subsection{Motion Representations and Measurement}
Pose estimators recover 2D joints~\cite{openpose2017,vitpose2022} and 3D motion from monocular video~\cite{videopose3d2019,gvhmr2024,wham2024}, and motion--language models learn shared representations of motion and text~\cite{motiongpt2023,motionllm2024,llamo2025,figmo2027,biomechgpt2025}.
Explicit approaches derive geometric measurements for fitness feedback, joint-angle monitoring, markerless biomechanics, and gesture analysis~\cite{aifit2021,trujillo2019,biocoach2026,agenticjointangle2026,opencap2023}, generate rule-based motion descriptions and question--answer pairs~\cite{posescript2022,motionscript2025}, prompt language models with computed kinematic attributes~\cite{kpmbench2026}, or give them structured, partly numerical descriptions of a sequence~\cite{smd2026}.
These approaches fix their measurements or descriptions for each sequence or task, whereas QMM computes measurements at question time for the videos, quantity, and interval that the VLM chooses, under definitions shared across videos.

\subsection{VLM Analysis, Programs, and Memory}
Benchmarks evaluate motion, temporal, and multi-video understanding~\cite{motionbench2025,temporalbench2024,videosetdiff2025,crossvid2026}, VidDiff and CoachMe compare action videos~\cite{viddiff2025,coachme2025}, and VLMs struggle with quantitative physical and motion judgments~\cite{quantiphy,motionhalluc}. VidDiff further identifies localizing the relevant sub-actions as a key failure of models that view frames, with little change across frame rates~\cite{viddiff2025}.
Temporal grounding localizes moments described in language~\cite{momentdetr2021,timelens2026}, whereas QMM retrieves the window of a requested duration that maximizes a measured quantity.
NSPose and IMoRe answer categorical questions about motion with programs over skeleton sequences~\cite{nspose2023,imore2025}, AmadeusGPT generates programs over animal pose keypoints through a behavior-analysis API~\cite{amadeusgpt2023}, ChatHuman and ChatMotion call human-centric and motion-analysis tools~\cite{chathuman2025,chatmotion2025}, and motion-grounded models reason over skeleton sequences or videos~\cite{mochat2024,motionsight2026}. Each of these answers questions about one image, motion, or video per request.
BiomechAgent writes code over stored multiview markerless kinematics, and domain formulas and validated gait tools improve its answers~\cite{biomechagent2026}. MotionHalluc's Perceive-Parse-Verify verifies corrective instructions for a video pair with kinematic measurements at key frames located by kinematic extrema~\cite{motionhalluc}.
Program-based methods compose visual operations~\cite{visprog2023,vipergpt,proviq2024}, and spatial reasoning methods train VLMs on metric 3D data~\cite{spatialvlm2024} or measure static scenes with programs and tools, an approach that outperforms judgments from pixels~\cite{vadar2025,seeingnotmeasuring2026}.
Long-video memories store content for reuse~\cite{worldmm,videoemem2026} or build it for each query, including tables of tracked objects~\cite{videoarm,doraemongpt2024}, and HumanSkill-4D concurrently applies shared geometric operators to stored avatar motion to select body regions and motion intervals on one avatar~\cite{humanskill4d2026}. Other concurrent training-free agents query a physics-fitted world built for each video to reason about objects~\cite{physmind2026} or call gait-measurement tools to diagnose one video~\cite{drgait2026}.
Unlike programs written for each question or operators on one avatar, QMM estimates a motion record from each ordinary monocular video once, reuses it across questions and collections, and fixes measurement definitions shared across videos, with selected videos and retrieved intervals restricting later measurements.

\section{Quantitative Motion Memory}\label{sec:method}

QMM separates what is computed once per video from what is computed for each question (Fig.~\ref{fig:framework}).

\subsection{Motion Records and Conventions}\label{sec:QMM}

\paragraph{Time-indexed motion records.}
For each RGB video $V_v$, an observer $\mathcal{O}$, any process that produces time-indexed estimates, builds a motion record $M_v=\mathcal{O}(V_v)$ of question-independent estimates. Here $\mathcal{O}$ is a monocular 3D pose estimator.
The operators need only estimates on which the definitions can be computed, which for human motion are 3D joints with an up direction. Other estimators can therefore replace the pose estimator, and with a second estimator the answers follow its records (Sec.~\ref{sec:across_video_results}).
The record stores estimated joint positions for every frame, their timestamps, and per-joint validity.

\paragraph{Shared measurement conventions.}
Conventions $\mathcal{P}$ fix how QMM computes every quantity in every video.
They specify the body-relative origin of each quantity, the body scale used for normalization, the temporal filter applied before differentiation, and the validity rule that decides when a measurement is unavailable.
Each measurement covers an interval, because questions concern periods of motion and per-frame monocular estimates are noisy. The validity rule requires a minimum duration of valid observations within the interval.
QMM computes body scales once per video and applies the same choices to every interval, video, and question (Sec.~\ref{sec:exp_setup}).

\subsection{Measurement and Temporal Operators}\label{sec:qmm_operators}

\paragraph{Measurement.}
For quantity $k$ over interval $I$ in video $v$, QMM computes
\begin{equation}
    \hat{z}_{v,k}(I)=m_k(M_v,I;\mathcal{P}),
    \label{eq:qmm_measurement}
\end{equation}
where $m_k$ performs the quantity-specific computation and temporal aggregation under $\mathcal{P}$.
We call the quantity formulas of $m_k$ together with $\mathcal{P}$ the measurement definitions. They are identical for every video, so $\hat{z}_{v,k}(I)$ has the same meaning for every video and interval.
For example, the mean hand-speed measurement computes wrist velocities relative to the shoulder midpoint from the timestamped observations, normalizes their magnitudes by the video's body scale, and averages valid values over $I$.
The interval can be the entire video, a specified interval, an interval returned by an earlier request, or an interval supplied with the input, such as an annotated action resolved by name in each video. QMM returns the value or reports it unavailable.

\paragraph{Window search.}
Window search, a temporal operator, returns an interval rather than a value and builds on the interval measurement in Eq.~\eqref{eq:qmm_measurement}.
For a quantity $k$ and a requested window duration $\Delta$, window search ranks every candidate window of that duration by its interval measurement,
\begin{equation}
    I^\star_{v,k}=\operatorname*{arg\,max}_{I\in\mathcal{I}_\Delta(v)} m_k(M_v,I;\mathcal{P}),
    \label{eq:qmm_search}
\end{equation}
where $\mathcal{I}_\Delta(v)$ denotes the windows of duration $\Delta$ that lie within video $v$ and start on a regular temporal grid, restricted to windows whose measurements are available.
Window search returns the start and end times of $I^\star_{v,k}$, the earliest in case of ties, together with its value, or reports the search as unavailable when no window qualifies.
Because the same interval operator scores every candidate, the search applies to every quantity and every duration that satisfies the validity rule, and it returns an interval in the same form as a requested one.

\subsection{VLM-Guided Analysis}\label{sec:qmm_analysis}

\paragraph{Requests and composition.}
The VLM answers a single question or a multi-step question, called a chain, whose steps build on earlier answers, through a sequence of requests $r_1,\dots,r_n$.
Each request $r_t=(o_t,k_t,\mathcal{S}_t)$ specifies an operator $o_t$ (measurement, or window search with a window duration), a quantity $k_t$, and a scope $\mathcal{S}_t$ given as a set of video and interval pairs (videos for window search).
The VLM receives the names of the available quantities and can query their definitions, so it maps the wording of a question to a quantity itself.
A scope can be stated in the question, resolved from a named action, or constructed from the results of earlier requests, so a video selected at one step or an interval retrieved at another determines where later operators are applied.
The VLM forms comparisons, rankings, and set selections from the returned values, with a calculator.

\paragraph{Division of roles.}
The VLM interprets the question, chooses operators, quantities, and scopes, and composes the returned values and intervals. It inspects RGB frames when a judgment requires visual interpretation, such as recognizing an activity. QMM performs every measurement and search on the records.
QMM trains neither the VLM nor the estimator. The operators implement the definitions once and are exposed as tools. A VLM given the definitions could also execute Eqs.~\eqref{eq:qmm_measurement} and~\eqref{eq:qmm_search} in its own code. This keeps the record and definitions of QMM but leaves their implementation to the backbone. The operators make the values independent of the backbone and need less input than code written by the VLM (Sec.~\ref{sec:why_qmm}).

\begin{table*}[!t]
\centering
\footnotesize
\setlength{\tabcolsep}{2.5pt}
\caption{
\textbf{QMM raises comparison and selection accuracy against dataset-3D references for every VLM.}
C, T, and S are pairwise accuracy, top-1 accuracy, and set Jaccard against dataset-3D references, and VidDiff is agreement with human annotations.
Mixed combines clips from all three datasets.
Questions per column are 64/48/48 (Mixed), 60 (HuMMan, MoVi), 96/40/96 (Fit3D), and 88 (VidDiff). Most + QMM rows coincide (Sec.~\ref{sec:across_video_results}). Bold marks the better method.
}
\label{tab:across_video}
\begin{tabular*}{\textwidth}{@{\extracolsep{\fill}}ll rrr rrr rrr rrr r@{}}
\toprule
& & \multicolumn{3}{c}{Mixed} & \multicolumn{3}{c}{HuMMan} & \multicolumn{3}{c}{MoVi} & \multicolumn{3}{c}{Fit3D} & \\
\cmidrule(lr){3-5}\cmidrule(lr){6-8}\cmidrule(lr){9-11}\cmidrule(lr){12-14}
Model & Method & C & T & S & C & T & S & C & T & S & C & T & S & VidDiff \\
\midrule
Qwen3.8-27B & Agent & 92 & 75 & 71 & 75 & 40 & 38 & 78 & 40 & 52 & 69 & 32 & 44 & 69 \\
& + QMM & \textbf{97} & \textbf{96} & \textbf{91} & \textbf{100} & \textbf{88} & \textbf{87} & \textbf{98} & \textbf{90} & \textbf{97} & \textbf{93} & \textbf{62} & \textbf{73} & \textbf{94} \\
\midrule
Gemma-4-31B & Agent & 92 & 77 & 77 & 80 & 45 & 51 & 75 & 37 & 52 & 69 & 35 & 50 & 88 \\
& + QMM & \textbf{97} & \textbf{96} & \textbf{91} & \textbf{100} & \textbf{88} & \textbf{87} & \textbf{98} & \textbf{90} & \textbf{97} & \textbf{93} & \textbf{62} & \textbf{73} & \textbf{94} \\
\midrule
Qwen3.5-9B & Agent & 86 & 69 & 72 & 53 & 33 & 13 & 63 & 35 & 24 & 57 & 45 & 18 & 60 \\
& + QMM & \textbf{97} & \textbf{96} & \textbf{91} & \textbf{100} & \textbf{88} & \textbf{87} & \textbf{98} & \textbf{90} & \textbf{93} & \textbf{93} & \textbf{62} & \textbf{73} & \textbf{94} \\
\midrule
Qwen3.5-4B & Agent & 89 & 60 & 75 & 63 & 40 & 31 & 72 & 38 & 27 & 62 & 38 & 27 & 53 \\
& + QMM & \textbf{97} & \textbf{96} & \textbf{89} & \textbf{100} & \textbf{88} & \textbf{87} & \textbf{98} & \textbf{88} & \textbf{96} & \textbf{93} & \textbf{62} & \textbf{72} & \textbf{94} \\
\midrule
Gemma-4-E4B & Agent & 67 & 60 & 51 & 48 & 28 & 24 & 43 & 13 & 24 & 57 & 20 & 13 & 44 \\
& + QMM & \textbf{97} & \textbf{96} & \textbf{79} & \textbf{98} & \textbf{88} & \textbf{70} & \textbf{97} & \textbf{90} & \textbf{82} & \textbf{94} & \textbf{62} & \textbf{48} & \textbf{92} \\
\midrule
GPT-6.1 Sol (low) & Agent & 92 & 81 & 87 & 88 & 63 & 60 & 87 & 53 & 73 & 78 & 48 & 56 & \textbf{95} \\
& + QMM & \textbf{97} & \textbf{96} & \textbf{91} & \textbf{100} & \textbf{88} & \textbf{87} & \textbf{98} & \textbf{90} & \textbf{97} & \textbf{93} & \textbf{62} & \textbf{73} & 94 \\
\midrule
GPT-6.1 Sol (high) & Agent & 92 & 83 & 87 & 90 & 75 & 67 & 87 & 58 & 74 & 83 & 55 & 61 & \textbf{94} \\
& + QMM & \textbf{97} & \textbf{96} & \textbf{91} & \textbf{100} & \textbf{88} & \textbf{87} & \textbf{98} & \textbf{90} & \textbf{97} & \textbf{93} & \textbf{62} & \textbf{73} & \textbf{94} \\
\midrule
Gemini 3.8 Flash & Agent & 84 & 75 & 73 & 77 & 65 & 53 & 78 & 47 & 54 & 64 & 35 & 36 & 91 \\
& + QMM & \textbf{97} & \textbf{96} & \textbf{91} & \textbf{100} & \textbf{88} & \textbf{87} & \textbf{98} & \textbf{90} & \textbf{97} & \textbf{93} & \textbf{62} & \textbf{73} & \textbf{94} \\
\bottomrule
\end{tabular*}
\end{table*}

\section{Experiments}

\subsection{Experimental Setup}\label{sec:exp_setup}

\paragraph{Datasets and their roles.}
HuMMan~\cite{humman2022} provides short clips of many people performing the same exercises, so answers cannot be inferred from the activity alone.
MoVi~\cite{movi2021} provides untrimmed 76--202-s videos in which one person performs several activities in sequence. We use its annotated action intervals for cross-video questions and omit them for interval comparison, window search, and chains.
Fit3D~\cite{aifit2021} provides fitness exercises recorded under a different capture setup.
Mixed-source collections place two-second clips from all three datasets in one candidate pool of six videos, two per source, with the correct answer balanced across sources so that the source does not reveal it; their cross-video questions omit stance width, whose reference joints differ across datasets, and chains use six-second clips.
VidDiff~\cite{viddiff2025} provides human judgments of differences between paired HuMMan videos, of which we use the 88 statements about the measured attributes.
TED talks from the TED Gesture dataset~\cite{tedgesture2019} provide in-the-wild clips of one speaker each, which we use for the human evaluation (Sec.~\ref{sec:across_video_results}).

\paragraph{Motion records and quantities.}
Records are estimated from monocular RGB with GVHMR~\cite{gvhmr2024}, whose training data include the MoVi motion capture through AMASS~\cite{amass2019} but not its videos. MoVi records match their references about as closely as HuMMan records (Suppl.~Sec.~E), and Fit3D lies outside this training data.
We measure mean hand speed, mean foot speed, peak hand height, mean stance width, and peak knee flexion, adapting kinematic measures from gesture and squat biomechanics research~\cite{trujillo2019,escamilla2001}.
Quantities defined per side average both sides, hand quantities are taken relative to the shoulder midpoint and foot quantities relative to the hip midpoint, stance width is the 3D distance between the ankles, and knee flexion is 180\textdegree{} minus the hip--knee--ankle angle.
Lengths and speeds are normalized by the median shoulder width of each video, knee flexion is in degrees, and window search uses a 0.5-s grid (details in Suppl.~Secs.~B and~C).

\paragraph{Reference answers.}
Reference answers apply the same definitions to the 3D motion of each dataset, independently of the RGB-based records. Giving a VLM without QMM these definitions as text does not close its gap (Secs.~\ref{sec:across_video_results} and~\ref{sec:temporal_results}).
Changing the filters, the center joint of the hand quantities, or the side averaging alters at most 8\% of the reference answers and windows on any dataset (Suppl.~Sec.~B).
Two checks are independent of our formulas. The first is the VidDiff annotations. The second is a human evaluation in which three raters who did not take part in this work judge cross-video comparisons, TED pairs, and windows within a video. They answer plain-language questions that state side averaging and body-relative measurement, without our formulas or method outputs (Suppl.~Sec.~A).

\paragraph{Compared methods.}
The main comparator, the Agent, is the same VLM without QMM, inspecting RGB frames and using a calculator, and + QMM (QMM in the text) additionally calls the operators.
Both share the backbone, questions, answer schemas, and interaction budget, including a pooled limit of $24N$ frames for $N$ videos, and each can request 2--16 frames at 672\,px from any interval of any video.
Because the operators read records from every frame while the budget limits the frames the VLM views, further conditions vary the Agent's frame budget, give whole videos at one frame per second, the full definitions as text, or text notes written from frames. On the same records, they give the joint trajectories in the prompt, let the VLM write code over the raw joints with or without the definitions, or coarsen the record. Measurement-only answers use the measured values without a VLM (Suppl.~Sec.~D).
We also evaluate the temporal-grounding model TimeLens-8B~\cite{timelens2026} on window search. Prior motion--language and paired-video methods neither compare values across a collection nor search untrimmed videos for windows by a measured quantity, so Sec.~\ref{sec:why_qmm} tests the components of QMM, including code over the same records as in code-writing agents~\cite{amadeusgpt2023,biomechagent2026}.

\paragraph{Backbones.}
We evaluate seven VLM backbones in eight settings, namely the open models Qwen3.8-27B, Qwen3.5-9B, Qwen3.5-4B~\cite{qwen35_2026,qwen38_2026}, Gemma-4-31B, and Gemma-4-E4B~\cite{gemma4_2026} with thinking disabled for both methods, and the commercial models GPT-6.1 Sol (Sol)~\cite{gpt61sol_2026} at low and high reasoning effort and Gemini 3.8 Flash~\cite{gemini38flash_2026} at low effort.
Higher reasoning effort raises the Sol Agent by at most 12 points (HuMMan top-1) and its window search from 38\% to 40\%. Enabling thinking for the open Agents raises MoVi window search from 12\% to 18\% for Qwen3.8-27B and leaves Qwen3.5-9B at 6\% (Suppl.~Sec.~D). Further conditions with Sol use high effort, except the frame sweep (low effort).

\paragraph{Evaluation.}
C reports pairwise comparison accuracy, T top-1 accuracy, and S the Jaccard index for selecting candidates whose value exceeds that of an anchor video.
In all but 45 items, the compared reference values differ by more than a near-tie margin, set for all datasets to $0.25$ times the interquartile range of the quantity over HuMMan (23\textdegree{} for knee flexion), and scoring only the other items changes no conclusion. Both methods receive identical items.
Scores are on a 0--100 scale, and missing answers score zero. Paired 95\% bootstrap intervals and further breakdowns are in Suppl.~Sec.~E.

\subsection{Cross-Video Quantitative Analysis}\label{sec:across_video_results}

\begin{figure}[t]
\centering
\includegraphics[width=\columnwidth]{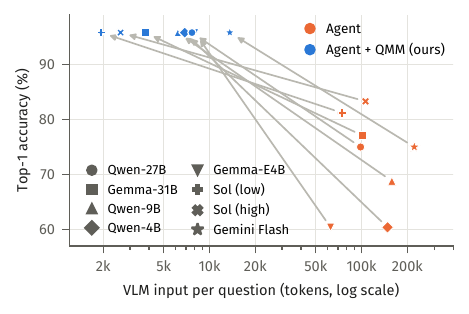}
\caption{
\textbf{QMM is more accurate with far less input.}
Top-1 accuracy versus VLM input tokens per question, excluding record construction, on the 48 mixed-source top-1 questions. Each arrow joins the Agent and + QMM of one setting, and every paired 95\% interval of the gain lies above zero.
}
\label{fig:mixed_cost}
\end{figure}

\paragraph{Comparison and selection.}
QMM improves cross-video comparison for every VLM. It scores higher than the Agent on comparison, top-1 selection, and set selection in all eight settings, both on mixed-source collections and on each dataset (Table~\ref{tab:across_video}), and 88 of the 96 paired-bootstrap intervals for C, T, and S lie above zero.
The gains are largest where the compared values differ least. On the 140 pairs whose reference values differ by less than twice the margin, QMM answers 94--95\% correctly and the Agents 49--82\%. Even for Sol at high effort, the gain is 13 points on these pairs against 5 points on the others, with all paired intervals above zero (Suppl.~Sec.~E).
Qwen3.5-4B with QMM scores higher than Sol without QMM at both reasoning efforts in every C, T, and S column. In all eight settings, QMM raises top-1 accuracy on mixed-source collections to 96\% with 8 to 41 times less input per question (Fig.~\ref{fig:mixed_cost}).
The gain results from computing the values rather than from coordinates or definitions alone. Given the joint trajectories of every video at 10\,Hz, Sol at high effort reaches 92/90/81 on C/T/S, and given the full definitions as text 94/92/86, against 97/96/91 with QMM.
With QMM, the VLMs differ little on these direct questions, where the VLM chooses the quantity and scope. Seven of the eight settings score within four points of one another in every C, T, and S column, and six are identical on mixed-source collections. Only Gemma-4-E4B scores lower, on set selection. VLMs that compute the values in their own code from the same records and definitions remain dependent on the backbone. So do composition, activity recognition, and quantities outside the operators (Secs.~\ref{sec:temporal_results} and~\ref{sec:why_qmm}).
Answers from frames also change more under rewording, and Agent scores vary by up to 10 points between runs while every QMM gain remains positive (Suppl.~Sec.~E).
As the collection grows, QMM remains far more accurate. At 64 candidate videos on HuMMan, the five open backbones reach 55--65\% top-1 with QMM against at most 20\% as Agents, and from 8 to 64 videos 39 of 40 paired intervals of the top-1 and set gains lie above zero.

\paragraph{Human judgments.}
The raters judge the 64 mixed-source comparisons of Table~\ref{tab:across_video} and 80 pairs of TED clips, paired without any measurement, that compare hand speed or hand height between speakers of different talks.
The raters read only the plain-language question, without our formulas. Their majority matches the dataset-3D reference on 50 of the 53 mixed-source items with a majority answer (94\%), including every hand-height, knee-flexion, and foot-speed item, so the references agree with what people read from the questions.
QMM agrees with the human majority on 96\% in every setting, against 70--96\% for the Agents. The raters are shown 16 close HuMMan and MoVi comparisons, including every close item on which QMM and the Sol Agent at high effort disagree, and give a majority answer on 9 of them. The majority agrees with QMM on all 9, including the 5 on which the two disagree, and with the Sol Agent on 4.
TED pairs have no 3D reference. On the 69 with a majority answer, QMM agrees with the majority on 90--91\% in every setting, as do measurement-only answers (91\%), against 61--88\% for the Agents. The paired intervals lie above zero for the four Agents below 80\%.

\paragraph{VidDiff annotations.}
On the 88 VidDiff statements, QMM agrees with 92--94\% of the human annotations in every setting, against 44--95\% for the Agents. It matches the Sol Agent, which agrees with 95\% and 94\% at low and high effort, and exceeds every other Agent.
On the 69 statements whose two videos differ by more than the near-tie margin on the records, QMM agrees with 99\% of the annotations in every setting except Gemma-4-E4B (96\%), as does the Sol Agent. Most remaining disagreements fall within the margin (Suppl.~Sec.~E).

\paragraph{Source of remaining errors.}
The remaining errors of QMM follow the motion records. On comparison and top-1 selection, its answers match the measurement-only answers within two points, and record values agree with the references at an intraclass correlation of 0.92--0.97 for every quantity except stance width.
With the operators fixed, measurement-only answers on mixed-source collections reach 97/96/91 on C/T/S with our records and 86/81/75 with records from a different pose estimator, MotionBERT~\cite{motionbert}. The same ordering holds on every dataset, so a stronger estimator improves the answers while the operators and definitions remain unchanged.

\subsection{Temporal and Multi-Step Analysis}\label{sec:temporal_results}

\begin{table}[t]
\centering
\footnotesize
\setlength{\tabcolsep}{2pt}
\caption{
\textbf{QMM raises window search and chains for every backbone.}
On MoVi, Int.\ compares two given intervals (120 questions), Win.\ finds the highest-valued 5-s window, correct when its start lies within $\pm$2.5\,s of the reference (120), and Chain compares another quantity over that window (121).
Mixed-source chains (41) follow Fig.~\ref{fig:framework}, and Quant., Act., and Full require all quantitative answers, the activity, and all answers. Bold marks the better method.
}
\label{tab:temporal}
\begin{tabular*}{\columnwidth}{@{\extracolsep{\fill}}ll rrr rrr@{}}
\toprule
& & \multicolumn{3}{c}{MoVi} & \multicolumn{3}{c}{Mixed-source chains} \\
\cmidrule(lr){3-5}\cmidrule(lr){6-8}
Model & Method & Int. & Win. & Chain & Quant. & Act. & Full \\
\midrule
Qwen3.8-27B & Agent & 81 & 12 & 14 & 22 & 44 & 20 \\
& + QMM & \textbf{99} & \textbf{98} & \textbf{98} & \textbf{95} & \textbf{90} & \textbf{90} \\
\midrule
Gemma-4-31B & Agent & 80 & 10 & 6 & 22 & 49 & 22 \\
& + QMM & \textbf{99} & \textbf{98} & \textbf{98} & \textbf{95} & \textbf{83} & \textbf{83} \\
\midrule
Qwen3.5-9B & Agent & 67 & 6 & 4 & 15 & 41 & 15 \\
& + QMM & \textbf{99} & \textbf{98} & \textbf{98} & \textbf{95} & \textbf{80} & \textbf{80} \\
\midrule
Qwen3.5-4B & Agent & 58 & 2 & 0 & 27 & 51 & 24 \\
& + QMM & \textbf{99} & \textbf{98} & \textbf{98} & \textbf{95} & \textbf{78} & \textbf{76} \\
\midrule
Gemma-4-E4B & Agent & 55 & 5 & 0 & 7 & 15 & 5 \\
& + QMM & \textbf{99} & \textbf{95} & \textbf{89} & \textbf{85} & \textbf{37} & \textbf{37} \\
\midrule
GPT-6.1 Sol (low) & Agent & 87 & 38 & 28 & 39 & 76 & 39 \\
& + QMM & \textbf{99} & \textbf{98} & \textbf{98} & \textbf{95} & \textbf{95} & \textbf{95} \\
\midrule
GPT-6.1 Sol (high) & Agent & 88 & 40 & 29 & 49 & 78 & 49 \\
& + QMM & \textbf{99} & \textbf{98} & \textbf{98} & \textbf{95} & \textbf{95} & \textbf{95} \\
\midrule
Gemini 3.8 Flash & Agent & 84 & 36 & 23 & 27 & 49 & 27 \\
& + QMM & \textbf{97} & \textbf{98} & \textbf{98} & \textbf{95} & \textbf{95} & \textbf{95} \\
\bottomrule
\end{tabular*}
\end{table}

\paragraph{Window search and chains.}
Searching the record finds intervals that sampled frames miss. On MoVi videos of 76--202 seconds without action intervals, QMM raises window search from 2--40\% to 95--98\% across the eight settings (Table~\ref{tab:temporal}). Averaged over the settings, it raises interval comparison from 75\% to 99\% and MoVi chains from 13\% to 97\%, and all 24 paired-bootstrap intervals lie above zero.
With QMM, the two smallest evaluated VLMs, Qwen3.5-4B and Gemma-4-E4B, reach 98\% and 95\% on window search, compared with 40\% for Sol at high effort without QMM. On the 41 mixed-source chains, QMM raises accuracy on the quantitative steps from 7--49\% to 85--95\%, and the same two VLMs reach 95\% and 85\% against 49\% for Sol. Complete chains rise from 15--49\% to 76--95\% in seven of the eight settings and from 5\% to 37\% for Gemma-4-E4B. In the same seven settings, QMM names the activity in the retrieved window correctly in 78--95\% of the chains against 41--78\% for the Agents, and with Gemma-4-E4B in 37\% against 15\%.
Fit3D videos are shorter and lie outside the training data of the pose estimator. On their 85 window searches, QMM reaches 93\% for all seven backbones against 21--52\% for the Agents, with chance at 32\% (Suppl.~Sec.~E).

\paragraph{Human judgments of windows.}
From the searches in which the Sol Agent at high effort or the Qwen3.8-27B Agent chooses a window more than 5\,s from the reference and the reference value is at least 1.5 times as large, 40 are drawn, and the raters compare the two windows from the plain-language question alone.
The majority selects the reference window in all 40 pairs, and QMM with the same backbone chooses it in 39. For 10 wrong Sol windows drawn without the value condition, the majority selects the reference in 7 of 8 judged pairs and QMM chooses it in all 8. The Agents thus miss differences that the raters see.

\paragraph{Alternatives without QMM.}
The gap persists when the Agent is given more frames, whole videos, or the definitions as text. Raising the budget of the Sol Agent at low effort from 24 to 96 frames per video improves window search from 38\% to 82\%. This is the highest accuracy of any condition without QMM, yet it uses 55 times the input of QMM with the same backbone and remains below its 98\%. With budgets of 144 and 192 frames it reaches 79\% and 77\% (Suppl.~Sec.~E).
With 96 frames or more, Sol inspects the correct window in over 90\% of the searches yet answers 82--87\% of those correctly.
Given whole videos at one frame per second, Gemini 3.8 Flash and Qwen3.8-27B reach 72\% and 58\%. Given the full definitions as text, the Agents reach 39\% for Sol and 3--7\% for Qwen3.8-27B and Qwen3.5-9B, no higher than without them.
TimeLens-8B, which localizes annotated actions at 42\% R1@0.5, finds 66\% of the MoVi and 44\% of the Fit3D windows against 95--98\% and 93\% for QMM (Suppl.~Sec.~E).

\subsection{What the VLM Needs}\label{sec:why_qmm}

\begin{table}[t]
\centering
\footnotesize
\setlength{\tabcolsep}{2.5pt}
\caption{
\textbf{Accuracy across backbones requires time-indexed records, shared definitions, and operators.}
(A) Components added to the Agent (first row) on the same records. C is accuracy on 64 mixed-source comparisons and Win.\ on 120 MoVi window searches. Rec.\ is the time-indexed joint coordinates, Def.\ the measurement definitions, and Op.\ the operators that implement them; without Op., the VLM implements the definitions in its own code.
(B) Stored record granularity on the MoVi questions of Table~\ref{tab:temporal} for Qwen3.8-27B calling measurement functions in code; peak moment keeps the time and value of the highest 1-s window. Bold marks the best.
}
\label{tab:design}
\begin{tabular*}{\columnwidth}{@{\extracolsep{\fill}}ccc cccccc@{}}
\toprule
\multicolumn{7}{@{}l}{\textit{\textbf{(A) Components}}} & \multicolumn{2}{c}{\makebox[0pt]{GPT-6.1}} \\
& & & \multicolumn{2}{c}{\makebox[0pt]{Qwen3.8-27B}} & \multicolumn{2}{c}{\makebox[0pt]{Qwen3.5-9B}} & \multicolumn{2}{c}{\makebox[0pt]{Sol (high)}} \\
\cmidrule(lr){4-5}\cmidrule(lr){6-7}\cmidrule(lr){8-9}
Rec. & Def. & Op. & C & Win. & C & Win. & C & Win. \\
\midrule
$\times$ & $\times$ & $\times$ & 92 & 12 & 86 & 6 & 92 & 40 \\
\checkmark & $\times$ & $\times$ & 80 & 62 & 84 & 58 & 91 & 81 \\
\checkmark & \checkmark & $\times$ & 91 & 91 & 77 & 67 & \textbf{97} & \textbf{98} \\
\checkmark & \checkmark & \checkmark & \textbf{97} & \textbf{98} & \textbf{97} & \textbf{98} & \textbf{97} & \textbf{98} \\
\bottomrule
\end{tabular*}

\vspace{3pt}
\begin{tabular*}{\columnwidth}{@{\extracolsep{\fill}}l rrr@{}}
\toprule
\multicolumn{4}{@{}l}{\textit{\textbf{(B) Record granularity}}} \\
Stored record & Int. & Win. & Chain \\
\midrule
Whole-video values & 78 & 7 & 5 \\
Peak moment & 72 & 66 & 45 \\
2.5-s summaries & 95 & 92 & 88 \\
1-s summaries & 97 & 97 & \textbf{95} \\
Time-indexed record & \textbf{98} & \textbf{98} & \textbf{95} \\
\bottomrule
\end{tabular*}

\end{table}

Table~\ref{tab:design}(A) adds the components of QMM one at a time, and its middle rows are ablation variants in which the VLM implements the definitions in its own code. The time-indexed record allows the VLM to locate intervals, shared definitions give each value the same meaning across videos, and operators remove the dependence on backbone-written code. RGB frames serve visual interpretation, such as recognizing activities. Without frames, tasks that require recognizing an activity lose 9--15 points on HuMMan for the two tested VLMs (Suppl.~Sec.~E).

\paragraph{Components.}
Each step uses the same questions, and every step with a record uses the same records.
The record allows the VLM to locate intervals, raising window search from 12\% to 62\% for Qwen3.8-27B and from 40\% to 81\% for Sol, whereas joint trajectories in the prompt at 2\,Hz give 12\% and 55\%.
Without the definitions, each program applies one computation to all videos but measures its own version of the quantity, such as hand speed without removing body motion or knee flexion as the interior angle (Suppl.~Sec.~E). Pairwise comparisons across videos are then no more accurate than from frames for any of the three backbones and 12 points lower for Qwen3.8-27B (80\% against 92\%).
Such programs also agree less with the human majority, at 83--94\% against 96\% for QMM, with the paired interval above zero for Qwen3.8-27B. Removing the hand-speed, knee-flexion, hand-height, and body-scale definitions together from the operators lowers accuracy on the 64 mixed-source comparisons from 97\% to 72\% (Suppl.~Sec.~E).
Sol's code matches the operators only when given the definitions, so even the strongest evaluated VLM needs both the records and the shared definitions. With them, the code of Qwen3.8-27B reaches 91\% on both tasks and that of Qwen3.5-9B 77\% on comparison.
The operators reach 97\% and 98\% for all three backbones with 2.7 and 4.7 times less window-search input than code with definitions from Qwen3.8-27B and Sol. The operators implement the definitions once for every backbone and question.

\paragraph{Record granularity.}
Whole-video values and stored peaks fall short (Table~\ref{tab:design}(B)). One stored value per video reaches 7\% on window search. Storing the peak moment of each quantity reaches 66\% on window search but 45\% on chains.
One-second summaries nearly match the full record, but they store only the five quantities and cannot answer the side-specific chains below, so the full record defers the choice of quantities to query time.

\paragraph{Paraphrased questions.}
An LLM outside the evaluated models paraphrased 48 questions of Tables~\ref{tab:across_video} and~\ref{tab:temporal} without measurement terms. With QMM, every open backbone requests the intended quantity and reaches 98\%, against 92--98\% with the template wording and 35--62\% for the Agents, so the gains hold beyond template wording (Suppl.~Sec.~E).

\paragraph{Quantities outside the operators.}
On 60 MoVi chains that search and compare side-specific joint quantities, such as the speed of the left ankle, Qwen3.8-27B and Qwen3.5-4B complete 95\% with joint-level tools. Gemma-4-E4B completes 27\% with these tools and 77\% in code. The record thus also serves quantities outside the five defined ones (Suppl.~Sec.~E).

\subsection{Query-Time Cost and Record Reuse}\label{sec:efficiency}

Once records are built, QMM answers 45 mixed-source questions and 20 MoVi window searches 2.7 to 8 times faster than the Agent for each open backbone. It views at most four frames per question on average against 16--75.

Building a record on one RTX A6000 takes 2.2--4.2\,s per second of video, and because the record serves every later question that contains the video, a two-minute record recovers its construction time after 14--31 window searches at the MoVi rate of 2.2\,s per second (Suppl.~Sec.~E).

\section{Conclusion}\label{sec:conclusion}

QMM is a general training-free design in which a VLM answers quantitative questions from values measured on an explicit, time-indexed record instead of judging them from frames. Perception estimates the record once from monocular RGB, shared definitions give each value the same meaning in every video and interval, and the VLM decides what to measure and views frames for visual judgments.
Instantiated for human motion, QMM enables VLMs of every evaluated size to compare movements across videos, locate intervals in untrimmed videos, and build on earlier answers. Their answers are more accurate against the dataset-3D references, agree with independent human judgments, and need much less input for questions across videos and less for window search. Their quantitative answers largely follow the records rather than the backbone.

\paragraph{Limitations and future work.}
Measurements inherit the errors of monocular motion estimation and of the body scale, which better estimators and confidence-aware validity can reduce without changing the operators or the definitions.
QMM is evaluated here for the motion of one person per video, in laboratory captures and in-the-wild TED talks.
Since only the observer and the definitions are specific to motion, natural next steps include further quantities such as tempo, multi-person video, and records from other observers such as object tracks.

{
    \small
    \bibliographystyle{ieeenat_fullname}
    \bibliography{main}

\begin{thebibliography}{67}
\providecommand{\natexlab}[1]{#1}
\providecommand{\url}[1]{\texttt{#1}}
\expandafter\ifx\csname urlstyle\endcsname\relax
  \providecommand{\doi}[1]{doi: #1}\else
  \providecommand{\doi}{doi: \begingroup \urlstyle{rm}\Url}\fi

\bibitem[Burgess et~al.(2025)Burgess, Wang, Zhang, Rau, Lozano, Dunlap,
  Darrell, and Yeung]{viddiff2025}
James Burgess, Xiaohan Wang, Yuhui Zhang, Anita Rau, Alejandro Lozano, Lisa
  Dunlap, Trevor Darrell, and Serena Yeung.
\newblock Video action differencing.
\newblock In \emph{International Conference on Learning Representations}, 2025.

\bibitem[Cai et~al.(2024)Cai, Tan, Zhang, Zou, Zhang, Yao, Zhu, Gu, Zhong,
  Shang, Dou, Park, Gao, Lee, and Yang]{temporalbench2024}
Mu Cai, Reuben Tan, Jianrui Zhang, Bocheng Zou, Kai Zhang, Feng Yao, Fangrui
  Zhu, Jing Gu, Yiwu Zhong, Yuzhang Shang, Yao Dou, Jaden Park, Jianfeng Gao,
  Yong~Jae Lee, and Jianwei Yang.
\newblock {TemporalBench}: Benchmarking fine-grained temporal understanding for
  multimodal video models.
\newblock arXiv:2410.10818, 2024.

\bibitem[Cai et~al.(2022)Cai, Ren, Zeng, Lin, Yu, Wang, Fan, Gao, Yu, Pan,
  Hong, Zhang, Loy, Yang, and Liu]{humman2022}
Zhongang Cai, Daxuan Ren, Ailing Zeng, Zhengyu Lin, Tao Yu, Wenjia Wang,
  Xiangyu Fan, Yang Gao, Yifan Yu, Liang Pan, Fangzhou Hong, Mingyuan Zhang,
  Chen~Change Loy, Lei Yang, and Ziwei Liu.
\newblock {HuMMan}: Multi-modal {4D} human dataset for versatile sensing and
  modeling.
\newblock In \emph{European Conference on Computer Vision}, pages 557--577,
  2022.

\bibitem[Cao et~al.(2017)Cao, Simon, Wei, and Sheikh]{openpose2017}
Zhe Cao, Tomas Simon, Shih-En Wei, and Yaser Sheikh.
\newblock Realtime multi-person {2D} pose estimation using part affinity
  fields.
\newblock In \emph{Proceedings of the IEEE Conference on Computer Vision and
  Pattern Recognition}, pages 7291--7299, 2017.

\bibitem[Chen et~al.(2024{\natexlab{a}})Chen, Xu, Kirmani, Ichter, Sadigh,
  Guibas, and Xia]{spatialvlm2024}
Boyuan Chen, Zhuo Xu, Sean Kirmani, Brian Ichter, Dorsa Sadigh, Leonidas
  Guibas, and Fei Xia.
\newblock {SpatialVLM}: Endowing vision-language models with spatial reasoning
  capabilities.
\newblock In \emph{Proceedings of the IEEE/CVF Conference on Computer Vision
  and Pattern Recognition}, pages 14455--14465, 2024{\natexlab{a}}.

\bibitem[Chen et~al.(2026)Chen, Wang, Wang, Wang, and Lu]{videoemem2026}
Kerui Chen, Jinglu Wang, Yun Wang, Xinjiang Wang, and Yan Lu.
\newblock Write once, query many: Evolving compact reusable memory for long
  videos.
\newblock arXiv:2610.24537, 2026.

\bibitem[Chen et~al.(2024{\natexlab{b}})Chen, Lu, Zeng, Zhang, Wang, Zhang, and
  Zhang]{motionllm2024}
Ling-Hao Chen, Shunlin Lu, Ailing Zeng, Hao Zhang, Benyou Wang, Ruimao Zhang,
  and Lei Zhang.
\newblock {MotionLLM}: Understanding human behaviors from human motions and
  videos.
\newblock arXiv:2405.20340, 2024{\natexlab{b}}.

\bibitem[Choudhury et~al.(2024)Choudhury, Niinuma, Kitani, and
  Jeni]{proviq2024}
Rohan Choudhury, Koichiro Niinuma, Kris~M. Kitani, and L{\'a}szl{\'o}~A. Jeni.
\newblock Video question answering with procedural programs.
\newblock In \emph{European Conference on Computer Vision}, 2024.

\bibitem[Cotton and Leonard(2026)]{biomechagent2026}
R.~James Cotton and Thomas Leonard.
\newblock {BiomechAgent}: {AI}-assisted biomechanical analysis through
  code-generating agents.
\newblock arXiv:2602.06975, 2026.

\bibitem[Delmas et~al.(2022)Delmas, Weinzaepfel, Lucas, Moreno-Noguer, and
  Rogez]{posescript2022}
Ginger Delmas, Philippe Weinzaepfel, Thomas Lucas, Francesc Moreno-Noguer, and
  Gr{\'e}gory Rogez.
\newblock {PoseScript}: {3D} human poses from natural language.
\newblock In \emph{European Conference on Computer Vision}, pages 346--362,
  2022.

\bibitem[Du et~al.(2026)Du, Fan, Nan, Xie, Zhou, Li, Yang, Yang, and
  Tai]{motionsight2026}
Yipeng Du, Tiehan Fan, Kepan Nan, Rui Xie, Penghao Zhou, Xiang Li, Jian Yang,
  Zhenheng Yang, and Ying Tai.
\newblock {MotionSight}: Boosting fine-grained motion understanding in
  multimodal {LLMs}.
\newblock In \emph{International Conference on Learning Representations}, 2026.

\bibitem[Endo et~al.(2023)Endo, Hsu, Li, and Wu]{nspose2023}
Mark Endo, Joy Hsu, Jiaman Li, and Jiajun Wu.
\newblock Motion question answering via modular motion programs.
\newblock In \emph{International Conference on Machine Learning}, pages
  9312--9328, 2023.

\bibitem[Escamilla et~al.(2001)Escamilla, Fleisig, Lowry, Barrentine, and
  Andrews]{escamilla2001}
Rafael~F. Escamilla, Glenn~S. Fleisig, Tracy~M. Lowry, Steven~W. Barrentine,
  and James~R. Andrews.
\newblock A three-dimensional biomechanical analysis of the squat during
  varying stance widths.
\newblock \emph{Medicine \& Science in Sports \& Exercise}, 33\penalty0
  (6):\penalty0 984--998, 2001.

\bibitem[Fan et~al.(2024)Fan, Ma, Wu, Du, Li, Gao, and Li]{videoagent_fan2024}
Yue Fan, Xiaojian Ma, Rujie Wu, Yuntao Du, Jiaqi Li, Zhi Gao, and Qing Li.
\newblock {VideoAgent}: A memory-augmented multimodal agent for video
  understanding.
\newblock In \emph{European Conference on Computer Vision}, 2024.

\bibitem[Fieraru et~al.(2021)Fieraru, Zanfir, Pirlea, Olaru, and
  Sminchisescu]{aifit2021}
Mihai Fieraru, Mihai Zanfir, Silviu~Cristian Pirlea, Vlad Olaru, and Cristian
  Sminchisescu.
\newblock {AIFit}: Automatic {3D} human-interpretable feedback models for
  fitness training.
\newblock In \emph{Proceedings of the IEEE/CVF Conference on Computer Vision
  and Pattern Recognition}, pages 9919--9928, 2021.

\bibitem[{Gemma Team}(2026)]{gemma4_2026}
{Gemma Team}.
\newblock Gemma 4 technical report.
\newblock arXiv:2607.02770, 2026.

\bibitem[Ghorbani et~al.(2021)Ghorbani, Mahdaviani, Thaler, Kording, Cook,
  Blohm, and Troje]{movi2021}
Saeed Ghorbani, Kimia Mahdaviani, Anne Thaler, Konrad Kording, Douglas~James
  Cook, Gunnar Blohm, and Nikolaus~F. Troje.
\newblock {MoVi}: A large multi-purpose human motion and video dataset.
\newblock \emph{PLOS ONE}, 16\penalty0 (6):\penalty0 e0253157, 2021.

\bibitem[Glantz and Grange(2026)]{seeingnotmeasuring2026}
Kai Glantz and Clemens Grange.
\newblock Seeing is not measuring: Tool-augmented metric spatial reasoning for
  vision-language models.
\newblock arXiv:2609.29073, 2026.

\bibitem[{Google DeepMind}(2026)]{gemini38flash_2026}
{Google DeepMind}.
\newblock {Gemini 3.8 Flash} model card.
\newblock API model \texttt{gemini-3.8-flash}, 2026.

\bibitem[Guo et~al.(2026)Guo, He, Mo, Xu, Liu, and Yu]{motionhalluc}
Weile Guo, Shenghong He, Danying Mo, Chengdong Xu, Xuexun Liu, and Chao Yu.
\newblock {MotionHalluc}: Diagnosing kinematic hallucinations in fine-grained
  motion reasoning.
\newblock arXiv:2606.23061, 2026.

\bibitem[Gupta and Kembhavi(2023)]{visprog2023}
Tanmay Gupta and Aniruddha Kembhavi.
\newblock Visual programming: Compositional visual reasoning without training.
\newblock In \emph{Proceedings of the IEEE/CVF Conference on Computer Vision
  and Pattern Recognition}, pages 14953--14962, 2023.

\bibitem[Hong et~al.(2025)Hong, Cheng, Yang, Wang, Wang, Gu, Huang, Dong, and
  Tang]{motionbench2025}
Wenyi Hong, Yean Cheng, Zhuoyi Yang, Weihan Wang, Lefan Wang, Xiaotao Gu, Shiyu
  Huang, Yuxiao Dong, and Jie Tang.
\newblock {MotionBench}: Benchmarking and improving fine-grained video motion
  understanding for vision language models.
\newblock In \emph{Proceedings of the IEEE/CVF Conference on Computer Vision
  and Pattern Recognition}, pages 8450--8460, 2025.

\bibitem[Ionescu et~al.(2014)Ionescu, Papava, Olaru, and
  Sminchisescu]{h36m2014}
Catalin Ionescu, Dragos Papava, Vlad Olaru, and Cristian Sminchisescu.
\newblock {Human3.6M}: Large scale datasets and predictive methods for {3D}
  human sensing in natural environments.
\newblock \emph{IEEE Transactions on Pattern Analysis and Machine
  Intelligence}, 36\penalty0 (7):\penalty0 1325--1339, 2014.

\bibitem[Ji et~al.(2026)Ji, Shen, Zhu, Kong, and Liu]{biocoach2026}
Yuyang Ji, Yixuan Shen, Shengjie Zhu, Yu Kong, and Feng Liu.
\newblock From {3D} pose to prose: Biomechanics-grounded vision--language
  coaching.
\newblock arXiv:2603.26938, 2026.

\bibitem[Jiang et~al.(2023)Jiang, Chen, Liu, Yu, Yu, and Chen]{motiongpt2023}
Biao Jiang, Xin Chen, Wen Liu, Jingyi Yu, Gang Yu, and Tao Chen.
\newblock {MotionGPT}: Human motion as a foreign language.
\newblock In \emph{Advances in Neural Information Processing Systems}, 2023.

\bibitem[Jiang et~al.(2024)Jiang, Xie, and Li]{rtmw2024}
Tao Jiang, Xinchen Xie, and Yining Li.
\newblock {RTMW}: Real-time multi-person {2D} and {3D} whole-body pose
  estimation.
\newblock arXiv:2407.08634, 2024.

\bibitem[Kim et~al.(2026)Kim, Kim, Han, and Baek]{humanskill4d2026}
Junsu Kim, Naeun Kim, Dongyoon Han, and Seungryul Baek.
\newblock {HumanSkill-4D}: Training-free language querying of animatable
  {Gaussian} avatars in space and time.
\newblock arXiv:2610.24452, 2026.

\bibitem[Lei et~al.(2021)Lei, Berg, and Bansal]{momentdetr2021}
Jie Lei, Tamara~L. Berg, and Mohit Bansal.
\newblock Detecting moments and highlights in videos via natural language
  queries.
\newblock In \emph{Advances in Neural Information Processing Systems}, 2021.

\bibitem[Li et~al.(2025{\natexlab{a}})Li, Sugandhika, Ee, Peh, Zhang, Yang,
  Rajan, and Fernando]{imore2025}
Chen Li, Chinthani Sugandhika, Yeo~Keat Ee, Eric Peh, Hao Zhang, Hong Yang,
  Deepu Rajan, and Basura Fernando.
\newblock {IMoRe}: Implicit program-guided reasoning for human motion {Q\&A}.
\newblock In \emph{Proceedings of the IEEE/CVF International Conference on
  Computer Vision}, pages 12987--12996, 2025{\natexlab{a}}.

\bibitem[Li et~al.(2026{\natexlab{a}})Li, Wang, Tan, Wang, Yan, Shi, Cai,
  Jiang, and Hu]{crossvid2026}
Jingyao Li, Jingyun Wang, Molin Tan, Haochen Wang, Cilin Yan, Likun Shi, Jiayin
  Cai, Xiaolong Jiang, and Yao Hu.
\newblock {CrossVid}: A comprehensive benchmark for evaluating cross-video
  reasoning in multimodal large language models.
\newblock In \emph{Proceedings of the AAAI Conference on Artificial
  Intelligence}, 2026{\natexlab{a}}.

\bibitem[Li et~al.(2025{\natexlab{b}})Li, Jia, Wang, An, Li, Hwang, and
  Belongie]{chatmotion2025}
Lei Li, Sen Jia, Jianhao Wang, Zhaochong An, Jiaang Li, Jenq-Neng Hwang, and
  Serge Belongie.
\newblock {ChatMotion}: A multimodal multi-agent for human motion analysis.
\newblock arXiv:2502.18180, 2025{\natexlab{b}}.

\bibitem[Li et~al.(2025{\natexlab{c}})Li, Jia, Wang, Jiang, Zhou, Dai, Zhang,
  Wu, and Hwang]{llamo2025}
Lei Li, Sen Jia, Jianhao Wang, Zhongyu Jiang, Feng Zhou, Ju Dai, Tianfang
  Zhang, Zongkai Wu, and Jenq-Neng Hwang.
\newblock Human motion instruction tuning.
\newblock In \emph{Proceedings of the IEEE/CVF Conference on Computer Vision
  and Pattern Recognition}, pages 17582--17591, 2025{\natexlab{c}}.

\bibitem[Li et~al.(2026{\natexlab{b}})Li, Xiang, Mao, Wei, Chen, Masood, Li,
  and Adeli]{quantiphy}
Puyin Li, Tiange Xiang, Ella Mao, Shirley Wei, Xinye Chen, Adnan Masood,
  Fei-Fei Li, and Ehsan Adeli.
\newblock {QuantiPhy}: A quantitative benchmark evaluating physical reasoning
  abilities of vision-language models.
\newblock In \emph{Proceedings of the IEEE/CVF Conference on Computer Vision
  and Pattern Recognition}, 2026{\natexlab{b}}.

\bibitem[Lin et~al.(2026)Lin, Zhu, Gong, Qin, and Wang]{kpmbench2026}
Boda Lin, Yongjie Zhu, Xiaocheng Gong, Wenyu Qin, and Meng Wang.
\newblock {KPM-Bench}: A kinematic parsing motion benchmark for fine-grained
  motion-centric video understanding.
\newblock arXiv:2602.17768, 2026.

\bibitem[Lin et~al.(2025)Lin, Feng, Liu, and Black]{chathuman2025}
Jing Lin, Yao Feng, Weiyang Liu, and Michael~J. Black.
\newblock {ChatHuman}: Chatting about {3D} humans with tools.
\newblock In \emph{Proceedings of the IEEE/CVF Conference on Computer Vision
  and Pattern Recognition}, pages 8150--8161, 2025.

\bibitem[Loper et~al.(2015)Loper, Mahmood, Romero, Pons-Moll, and
  Black]{smpl2015}
Matthew Loper, Naureen Mahmood, Javier Romero, Gerard Pons-Moll, and Michael~J.
  Black.
\newblock {SMPL}: A skinned multi-person linear model.
\newblock \emph{ACM Transactions on Graphics}, 34\penalty0 (6):\penalty0
  248:1--248:16, 2015.

\bibitem[Mahmood et~al.(2019)Mahmood, Ghorbani, Troje, Pons-Moll, and
  Black]{amass2019}
Naureen Mahmood, Nima Ghorbani, Nikolaus~F. Troje, Gerard Pons-Moll, and
  Michael~J. Black.
\newblock {AMASS}: Archive of motion capture as surface shapes.
\newblock In \emph{Proceedings of the IEEE/CVF International Conference on
  Computer Vision}, pages 5442--5451, 2019.

\bibitem[Markhorst et~al.(2026)Markhorst, Lin, Chew, van Gemert, and
  Zhang]{figmo2027}
Thomas Markhorst, Zhi-Yi Lin, Jouh~Yeong Chew, Jan van Gemert, and Xucong
  Zhang.
\newblock Fine-grained human motion understanding with language models.
\newblock arXiv:2606.20888, 2026.

\bibitem[Marsili et~al.(2025)Marsili, Agrawal, Yue, and Gkioxari]{vadar2025}
Damiano Marsili, Rohun Agrawal, Yisong Yue, and Georgia Gkioxari.
\newblock Visual agentic {AI} for spatial reasoning with a dynamic {API}.
\newblock In \emph{Proceedings of the IEEE/CVF Conference on Computer Vision
  and Pattern Recognition}, pages 19446--19455, 2025.

\bibitem[Mo et~al.(2024)Mo, Chen, Lin, Ni, Zeng, Hu, and Li]{mochat2024}
Jiawei Mo, Yixuan Chen, Rifen Lin, Yongkang Ni, Min Zeng, Xiping Hu, and Min
  Li.
\newblock {MoChat}: Joints-grouped spatio-temporal grounding {LLM} for
  multi-turn motion comprehension and description.
\newblock arXiv:2410.11404, 2024.

\bibitem[{OpenAI}(2026)]{gpt61sol_2026}
{OpenAI}.
\newblock {GPT-6.1 Sol}.
\newblock API model \texttt{gpt-6.1-sol}; model documentation, 2026.

\bibitem[Pavlakos et~al.(2019)Pavlakos, Choutas, Ghorbani, Bolkart, Osman,
  Tzionas, and Black]{smplx2019}
Georgios Pavlakos, Vasileios Choutas, Nima Ghorbani, Timo Bolkart, Ahmed A.~A.
  Osman, Dimitrios Tzionas, and Michael~J. Black.
\newblock Expressive body capture: {3D} hands, face, and body from a single
  image.
\newblock In \emph{Proceedings of the IEEE/CVF Conference on Computer Vision
  and Pattern Recognition}, pages 10975--10985, 2019.

\bibitem[Pavllo et~al.(2019)Pavllo, Feichtenhofer, Grangier, and
  Auli]{videopose3d2019}
Dario Pavllo, Christoph Feichtenhofer, David Grangier, and Michael Auli.
\newblock {3D} human pose estimation in video with temporal convolutions and
  semi-supervised training.
\newblock In \emph{Proceedings of the IEEE/CVF Conference on Computer Vision
  and Pattern Recognition}, pages 7753--7762, 2019.

\bibitem[Qiu et~al.(2025)Qiu, Sun, Yagi, Egami, Miyata, Fukuda, Hara, and
  Sagawa]{videosetdiff2025}
Yue Qiu, Yanjun Sun, Takuma Yagi, Shusaku Egami, Natsuki Miyata, Ken Fukuda,
  Kensho Hara, and Ryusuke Sagawa.
\newblock {VideoSetDiff}: Identifying and reasoning similarities and
  differences in similar videos.
\newblock In \emph{Proceedings of the IEEE/CVF International Conference on
  Computer Vision}, pages 12242--12252, 2025.

\bibitem[{Qwen Team}(2026{\natexlab{a}})]{qwen35_2026}
{Qwen Team}.
\newblock Qwen3.5: Towards native multimodal agents.
\newblock Model release; checkpoints \texttt{Qwen/Qwen3.5-9B} and
  \texttt{Qwen/Qwen3.5-4B}, 2026{\natexlab{a}}.

\bibitem[{Qwen Team}(2026{\natexlab{b}})]{qwen38_2026}
{Qwen Team}.
\newblock {Qwen3.8-27B}.
\newblock Model card; checkpoint \texttt{Qwen/Qwen3.8-27B}, 2026{\natexlab{b}}.

\bibitem[Shen et~al.(2024)Shen, Pi, Xia, Cen, Peng, Hu, Bao, Hu, and
  Zhou]{gvhmr2024}
Zehong Shen, Huaijin Pi, Yan Xia, Zhi Cen, Sida Peng, Zechen Hu, Hujun Bao,
  Ruizhen Hu, and Xiaowei Zhou.
\newblock World-grounded human motion recovery via gravity-view coordinates.
\newblock In \emph{SIGGRAPH Asia 2024 Conference Papers}, 2024.

\bibitem[Shin et~al.(2024)Shin, Kim, Halilaj, and Black]{wham2024}
Soyong Shin, Juyong Kim, Eni Halilaj, and Michael~J. Black.
\newblock {WHAM}: Reconstructing world-grounded humans with accurate {3D}
  motion.
\newblock In \emph{Proceedings of the IEEE/CVF Conference on Computer Vision
  and Pattern Recognition}, pages 2070--2080, 2024.

\bibitem[Sur{\'i}s et~al.(2023)Sur{\'i}s, Menon, and Vondrick]{vipergpt}
D\'idac Sur{\'i}s, Sachit Menon, and Carl Vondrick.
\newblock {ViperGPT}: Visual inference via {Python} execution for reasoning.
\newblock In \emph{Proceedings of the IEEE/CVF International Conference on
  Computer Vision}, pages 11888--11898, 2023.

\bibitem[Trujillo et~al.(2019)Trujillo, Vaitonyte, Simanova, and
  {\"O}zy{\"u}rek]{trujillo2019}
James~P. Trujillo, Julija Vaitonyte, Irina Simanova, and Asli {\"O}zy{\"u}rek.
\newblock Toward the markerless and automatic analysis of kinematic features: A
  toolkit for gesture and movement research.
\newblock \emph{Behavior Research Methods}, 51:\penalty0 769--777, 2019.

\bibitem[Uhlrich et~al.(2023)Uhlrich, Falisse, Kidzi{\'n}ski, Muccini, Ko,
  Chaudhari, Hicks, and Delp]{opencap2023}
Scott~D. Uhlrich, Antoine Falisse, {\L}ukasz Kidzi{\'n}ski, Julie Muccini,
  Michael Ko, Akshay~S. Chaudhari, Jennifer~L. Hicks, and Scott~L. Delp.
\newblock {OpenCap}: Human movement dynamics from smartphone videos.
\newblock \emph{PLOS Computational Biology}, 19\penalty0 (10):\penalty0
  e1011462, 2023.

\bibitem[Wang et~al.(2024)Wang, Zhang, Zohar, and
  Yeung-Levy]{videoagent_wang2024}
Xiaohan Wang, Yuhui Zhang, Orr Zohar, and Serena Yeung-Levy.
\newblock {VideoAgent}: Long-form video understanding with large language model
  as agent.
\newblock In \emph{European Conference on Computer Vision}, 2024.

\bibitem[Xu et~al.(2022)Xu, Zhang, Zhang, and Tao]{vitpose2022}
Yufei Xu, Jing Zhang, Qiming Zhang, and Dacheng Tao.
\newblock {ViTPose}: Simple vision transformer baselines for human pose
  estimation.
\newblock In \emph{Advances in Neural Information Processing Systems}, 2022.

\bibitem[Yang et~al.(2026)Yang, Zeng, Zhao, Xu, He, Li, Deng, Fan, and
  Wang]{physmind2026}
Chen Yang, Shenxiang Zeng, Haoyang Zhao, Zhouyuan Xu, Youquan He, Haoyu Li,
  Mingyi Deng, Jiansheng Fan, and Chen Wang.
\newblock {PhysMind}: From video to executable worlds for training-free
  physical reasoning.
\newblock arXiv:2608.04575, 2026.

\bibitem[Yang et~al.(2025)Yang, Kennedy, and Cotton]{biomechgpt2025}
Ruize Yang, Ann Kennedy, and R.~James Cotton.
\newblock {BiomechGPT}: Extending motion-language models to clinical motion
  understanding.
\newblock arXiv:2505.18465, 2025.

\bibitem[Yang et~al.(2024)Yang, Chen, Li, Wang, and Yang]{doraemongpt2024}
Zongxin Yang, Guikun Chen, Xiaodi Li, Wenguan Wang, and Yi Yang.
\newblock {DoraemonGPT}: Toward understanding dynamic scenes with large
  language models (exemplified as a video agent).
\newblock In \emph{International Conference on Machine Learning}, 2024.

\bibitem[Yazdian et~al.(2025)Yazdian, Lagasse, Mohammadi, Liu, Cheng, and
  Lim]{motionscript2025}
Payam~Jome Yazdian, Rachel Lagasse, Hamid Mohammadi, Eric Liu, Li Cheng, and
  Angelica Lim.
\newblock {MotionScript}: Natural language descriptions for expressive {3D}
  human motions.
\newblock In \emph{IEEE/RSJ International Conference on Intelligent Robots and
  Systems}, 2025.

\bibitem[Ye et~al.(2023)Ye, Lauer, Zhou, Mathis, and Mathis]{amadeusgpt2023}
Shaokai Ye, Jessy Lauer, Mu Zhou, Alexander Mathis, and Mackenzie~W. Mathis.
\newblock {AmadeusGPT}: A natural language interface for interactive animal
  behavioral analysis.
\newblock In \emph{Advances in Neural Information Processing Systems}, 2023.

\bibitem[Yeh et~al.(2025)Yeh, Su, Chen, Lin, Ku, Chiu, Hu, and Ku]{coachme2025}
Wei-Hsin Yeh, Yu-An Su, Chih-Ning Chen, Yi-Hsueh Lin, Calvin Ku, Wen-Hsin Chiu,
  Min-Chun Hu, and Lun-Wei Ku.
\newblock {CoachMe}: Decoding sport elements with a reference-based coaching
  instruction generation model.
\newblock In \emph{Proceedings of the Annual Meeting of the Association for
  Computational Linguistics}, pages 29126--29151, 2025.

\bibitem[Yeo et~al.(2026)Yeo, Kim, Yoon, and Hwang]{worldmm}
Woongyeong Yeo, Kangsan Kim, Jaehong Yoon, and Sung~Ju Hwang.
\newblock {WorldMM}: Dynamic multimodal memory agent for long video reasoning.
\newblock In \emph{Proceedings of the IEEE/CVF Conference on Computer Vision
  and Pattern Recognition}, 2026.

\bibitem[Yin et~al.(2026{\natexlab{a}})Yin, Wang, Alamri, Aljohani, Liu,
  Fiedler, and Gao]{drgait2026}
Xiangyu Yin, Shiqi Wang, Abrar Alamri, Yasir Aljohani, Weichen Liu, Goeran
  Fiedler, and Wei Gao.
\newblock {DrGait}: Biomechanically grounded visual reasoning for interpretable
  clinical gait analysis.
\newblock arXiv:2609.28796, 2026{\natexlab{a}}.

\bibitem[Yin et~al.(2026{\natexlab{b}})Yin, Meng, Chen, Ding, Shao, and
  Yu]{videoarm}
Yufei Yin, Qianke Meng, Minghao Chen, Jiajun Ding, Zhenwei Shao, and Zhou Yu.
\newblock {VideoARM}: Agentic reasoning over hierarchical memory for long-form
  video understanding.
\newblock In \emph{Proceedings of the IEEE/CVF Conference on Computer Vision
  and Pattern Recognition}, 2026{\natexlab{b}}.

\bibitem[Yoon et~al.(2019)Yoon, Ko, Jang, Lee, Kim, and Lee]{tedgesture2019}
Youngwoo Yoon, Woo-Ri Ko, Minsu Jang, Jaeyeon Lee, Jaehong Kim, and Geehyuk
  Lee.
\newblock Robots learn social skills: End-to-end learning of co-speech gesture
  generation for humanoid robots.
\newblock In \emph{IEEE International Conference on Robotics and Automation},
  2019.

\bibitem[Yu et~al.(2026)Yu, A, Xie, and Wang]{agenticjointangle2026}
Juncheng Yu, Lusi A, Haoxuan Xie, and Weiming Wang.
\newblock Agentic pipeline for self-synchronized multiview joint angle
  monitoring in uncalibrated environments.
\newblock In \emph{Proceedings of the International Conference of the IEEE
  Engineering in Medicine and Biology Society}, 2026.

\bibitem[Zhang et~al.(2026{\natexlab{a}})Zhang, Wang, Ge, Ge, Li, Shan, and
  Wang]{timelens2026}
Jun Zhang, Teng Wang, Yuying Ge, Yixiao Ge, Xinhao Li, Ying Shan, and Limin
  Wang.
\newblock {TimeLens}: Rethinking video temporal grounding with multimodal
  {LLMs}.
\newblock In \emph{Proceedings of the IEEE/CVF Conference on Computer Vision
  and Pattern Recognition}, 2026{\natexlab{a}}.

\bibitem[Zhang et~al.(2026{\natexlab{b}})Zhang, Liu, Ploetz, and Xiao]{smd2026}
Yao Zhang, Zhuchenyang Liu, Thomas Ploetz, and Yu Xiao.
\newblock Encoder-free human motion understanding via structured motion
  descriptions.
\newblock arXiv:2604.21668, 2026{\natexlab{b}}.

\bibitem[Zhu et~al.(2023)Zhu, Ma, Liu, Liu, Wu, and Wang]{motionbert}
Wentao Zhu, Xiaoxuan Ma, Zhaoyang Liu, Libin Liu, Wayne Wu, and Yizhou Wang.
\newblock {MotionBERT}: A unified perspective on learning human motion
  representations.
\newblock In \emph{Proceedings of the IEEE/CVF International Conference on
  Computer Vision}, pages 15085--15099, 2023.

\end{thebibliography}
}

% supplement float placement (arXiv layout only)
\renewcommand{\topfraction}{0.9}\renewcommand{\dbltopfraction}{0.9}\renewcommand{\bottomfraction}{0.8}
\renewcommand{\textfraction}{0.08}\renewcommand{\floatpagefraction}{0.85}\renewcommand{\dblfloatpagefraction}{0.85}
\setcounter{topnumber}{4}\setcounter{bottomnumber}{3}\setcounter{totalnumber}{6}\setcounter{dbltopnumber}{3}
\makeatletter\setlength{\@fptop}{0pt}\setlength{\@dblfptop}{0pt}\makeatother
\raggedbottom
\clearpage
\maketitlesupplementary

\renewcommand{\thesection}{\Alph{section}}
\setcounter{section}{0}
\providecommand{\theHsection}{}\renewcommand{\theHsection}{S\Alph{section}}
\setcounter{table}{0}\renewcommand{\thetable}{S\arabic{table}}\providecommand{\theHtable}{}\renewcommand{\theHtable}{S\arabic{table}}
\setcounter{figure}{0}\renewcommand{\thefigure}{S\arabic{figure}}\providecommand{\theHfigure}{}\renewcommand{\theHfigure}{S\arabic{figure}}

This supplement gives the evaluated tasks, example questions, and human evaluation (Sec.~\ref{sec:supp_tasks}), the measurement definitions and reference answers (Sec.~\ref{sec:supp_refs}), implementation details of QMM (Sec.~\ref{sec:supp_qmm}), the settings of the compared methods (Sec.~\ref{sec:supp_settings}), additional results in the order of Secs.~4.2--4.5 (Sec.~\ref{sec:supp_results}), a worked example (Sec.~\ref{sec:supp_example}), and the prompts (Sec.~\ref{sec:supp_prompts}). Most tables abbreviate the backbones as Qwen-27B (Qwen3.8-27B), Gemma-31B (Gemma-4-31B), Qwen-9B (Qwen3.5-9B), Qwen-4B (Qwen3.5-4B), Gemma-E4B (Gemma-4-E4B), Sol (GPT-6.1 Sol), and Gemini Flash (Gemini 3.8 Flash).

\section{Tasks, Questions, and Human Evaluation (Secs.~4.1--4.3)}\label{sec:supp_tasks}

\begin{table*}[t]
\centering
\scriptsize
\setlength{\tabcolsep}{3pt}
\caption{\textbf{Item sets of the human evaluation}, each judged by the same three raters. Indented rows are additional items of the same type as the set above.}
\label{tab:supp_humansets}
\begin{tabular*}{\textwidth}{@{\extracolsep{\fill}}l r >{\raggedright\arraybackslash}p{8.6cm} >{\raggedright\arraybackslash}p{3.4cm}@{}}
\toprule
Set & Items & Selection & Clips \\
\midrule
1. Cross-video, laboratory & 64 & All mixed-source comparisons of Table~1, 16 per quantity & The 2-s clips of Table~1 \\
\addlinespace[2pt]
\quad\textit{close comparisons} & 16 & HuMMan and MoVi comparisons of Table~1 whose reference values differ by less than twice the margin, namely all 9 on which the Sol Agent at high effort and QMM disagree, and 7 more drawn with a fixed seed over the quantities & The compared clips of Table~1 \\
\addlinespace
2. Cross-video, TED & 80 & 40 pairs per quantity for hand speed and hand height, the two that upper-body shots show, from 222 clips of 67 talks with one speaker per shot; the two clips of a pair come from different talks, each clip is used at most once per quantity, and a fixed seed draws the pairs without any measurement, model output, or human answer, and the 80 pairs span 51 talks & TED clips of 5--21\,s \\
\addlinespace
3. Windows in a video & 40 & Searches of Table~2 in which the Sol Agent at high effort (20) or the Qwen3.8-27B Agent (20) answered a window starting more than 5\,s from the reference and the reference value is at least 1.5 times that of the Agent's window (for the signed hand height, half a shoulder width higher), which holds for 54 of the 72 and 79 of the 89 such windows; fixed seed, balanced over quantities as far as available (10 hand speed, 10 foot speed, 9 hand height, 8 stance width, 3 knee flexion) & The two 5-s windows of the same MoVi video \\
\addlinespace[2pt]
\quad\textit{random wrong windows} & 10 & Drawn with a fixed seed from the other wrong windows of the Sol Agent at high effort that start more than 5\,s from the reference, without the value condition & As above \\
\bottomrule
\end{tabular*}
\end{table*}

\begin{table}[t]
\centering
\footnotesize
\setlength{\tabcolsep}{3pt}
\caption{\textbf{Agreement with human judgments.} Accuracy against the human majority on the judged items (53 mixed-source comparisons, 69 TED pairs), and for the mixed-source items also against the dataset-3D reference. Agent / QMM.}
\label{tab:supp_human}
\begin{tabular*}{\columnwidth}{@{\extracolsep{\fill}}l ccc@{}}
\toprule
& \multicolumn{2}{c}{Mixed} & TED \\
\cmidrule(lr){2-3}\cmidrule(lr){4-4}
Model & Human & Reference & Human \\
\midrule
Qwen-27B & 96 / 96 & 94 / 98 & 87 / 90 \\
Gemma-31B & 96 / 96 & 94 / 98 & 70 / 91 \\
Qwen-9B & 91 / 96 & 92 / 98 & 75 / 91 \\
Qwen-4B & 91 / 96 & 92 / 98 & 75 / 91 \\
Gemma-E4B & 70 / 96 & 68 / 98 & 61 / 90 \\
Sol (low) & 96 / 96 & 94 / 98 & 88 / 91 \\
Sol (high) & 96 / 96 & 94 / 98 & 88 / 91 \\
Gemini Flash & 94 / 96 & 89 / 98 & 88 / 91 \\
\bottomrule
\end{tabular*}
\end{table}

\subsection{Question Sets and Scoring}

\paragraph{Questions and counts.}
Questions are generated from templates with one fixed phrase per quantity and task, and both methods receive the same question text.
Table~\ref{tab:supp_counts} gives the scored questions, and Tab.~\ref{tab:supp_examples} gives one question per task with its answer and the reference values that determine it.

\begin{table*}[t]
\centering
\footnotesize
\setlength{\tabcolsep}{3pt}
\caption{\textbf{Evaluated questions.} Durations are clip lengths for HuMMan, Fit3D, and mixed-source collections, video lengths for MoVi, and frame counts at 8\,fps for VidDiff. Chains are the 121 MoVi chains with an unambiguous reference window, and cross-video questions on mixed-source collections do not include stance width, whose reference joints differ across datasets (Sec.~\ref{sec:supp_refs}).}
\label{tab:supp_counts}
\begin{tabular*}{\textwidth}{@{\extracolsep{\fill}}l l r r l l@{}}
\toprule
Task & Data & Questions & Candidates & Duration & Quantities \\
\midrule
C / T / S & Mixed & 64 / 48 / 48 & 6 (2 per source) & 2\,s & four (no stance width) \\
C / T / S & HuMMan & 60 / 60 / 60 & 6 (same exercise) & 2.1--13.5\,s & all five \\
C / T / S & MoVi (action intervals) & 60 / 60 / 60 & 8 & 76--202\,s & all five \\
C / T / S & Fit3D & 96 / 40 / 96 & 8 (same exercise) & 10.4--54.7\,s & all five \\
Comparison & VidDiff & 88 & 2 & 8--27 frames & stance, knee \\
Int. / Win. & MoVi & 120 / 120 & 1 video & 76--202\,s & all five \\
Chain & MoVi & 121 & 1 video & 76--198\,s & all five \\
Mixed chain & Mixed & 41 (3 turns) & 6 (2 per source) & 6\,s & hand speed, hand height, stance \\
\bottomrule
\end{tabular*}
\end{table*}

\begin{table*}[t]
\centering
\footnotesize
\setlength{\tabcolsep}{3pt}
\caption{\textbf{Example questions.} Reference values are computed from dataset 3D motion with the definitions of Tab.~\ref{tab:supp_quantities}. Fig.~2 and Sec.~\ref{sec:supp_example} give a mixed-source chain.}
\label{tab:supp_examples}
\begin{tabular*}{\textwidth}{@{\extracolsep{\fill}}l >{\raggedright\arraybackslash}p{0.55\textwidth} >{\raggedright\arraybackslash}p{0.3\textwidth}@{}}
\toprule
Task & Question & Answer and reference values \\
\midrule
C (mixed) & In which of V05 and V06 does the person raise the hands higher (relative to their own body size) over the whole clip? Answer V05 or V06. & V05. Hand height V05 $-0.05$, V06 $-0.87$ \\
\addlinespace[2pt]
T (mixed) & In which single video does the person move the hands the fastest (relative to their own body size) over the whole clip? & V05. Hand speed 2.62 vs.\ at most 0.86 for the others \\
\addlinespace[2pt]
S (mixed) & In which of the other videos does the person raise the hands higher than in V01 (relative to their own body size) over the whole clip? Do not include V01. & \{V03\}. V01 0.46, V03 1.60, others $\le 0.18$ \\
\addlinespace[2pt]
Int. & In V01, compare interval A (72.0--82.0\,s) and interval B (2.9--12.9\,s): during which interval does the person stand with the feet wider apart (relative to their own body size)? Answer A or B. & A. Stance width 0.83 vs.\ 0.56 \\
\addlinespace[2pt]
Win. & In V01, during which 5-second window does the person move the hands the fastest (relative to their own body size)? Give the start time of that window in seconds. & 5.5\,s (3.0--8.0 accepted). 6.24 vs.\ 3.33 for the best window starting more than 5\,s away \\
\addlinespace[2pt]
Chain & (1) In V01, during which 5-second window does the person stand with the feet widest apart (relative to their own body size)? Give the start time of that window in seconds. (2) Keep the window you just found as interval A, and take 105.0--110.0\,s of V01 as interval B: during which interval does the person raise the hands higher (relative to their own body size)? Answer A or B. & (1) 121.5\,s. (2) B. Hand height A $-1.08$, B $-0.50$ \\
\addlinespace[2pt]
VidDiff & Both videos show the exercise `Hip circle anticlockwise' (fitness exercise called standing hip circle with hands on hips, one rotation anticlockwise). In which video is the following more true: the feet stance is wider? Answer V01 or V02. & V02 (human annotation) \\
\bottomrule
\end{tabular*}
\end{table*}

\paragraph{Collections and chains.}
Mixed-source collections draw the central 2-s window of HuMMan clips, MoVi action segments, and Fit3D repetitions, with two clips per source, six distinct people and videos, and shuffled labels.
Pairs in C always come from different sources, top-1 winners and set anchors are balanced across sources, and C winners are balanced against the median order of the sources.
MoVi chains take the window found in step 1 as interval A and a fixed 5-s interval B more than 5\,s away whose step-2 value is closest to the median, with the two values differing by more than the near-tie margin $\delta$ (Sec.~\ref{sec:supp_refs}).
Chains are scored when the step-1 window is unique and the step-2 values differ by more than $\delta$.
Mixed-source chains use six 6-s clips, two per source, select a video by hand speed, search that video for the 2-s window with the highest hand height and name its activity, and compare stance width between that window and a second interval of the same video.
They require that the highest 2-s window over all six videos lies in another video, so that ignoring the first answer leads to a wrong second answer.

\paragraph{VidDiff.}
We use the closed statements of the VidDiffBench fitness domain whose categories state an implemented quantity, namely wider stance (stance width, frontal views) and less knee bend (knee flexion, reversed), which gives 88 statements over four actions (Tab.~\ref{tab:supp_viddiff}).

\paragraph{Scoring.}
Chance accuracy is 50\% for C, Int., and VidDiff, 17\% or 13\% for T with six or eight candidates, and 3--7\% for Win.\ given the video lengths.
C, T, Int., and VidDiff score 1 when the answered label matches, and S scores the Jaccard index between the answered and correct sets over the scored candidates.
Win.\ accepts a start within 2.5\,s of the correct 5-s window, and a MoVi chain scores 1 only when both steps are correct.
In mixed chains, the window step requires the correct video and a start within 1\,s of the correct 2-s window, Quant.\ requires all three quantitative answers, Act.\ the activity label, and Full all of them.

\paragraph{Data use.}
The datasets are used for non-commercial research under their terms, and the source videos are not redistributed; the questions and reference answers that we will release refer to the original datasets, which users obtain under their licenses. Frames of the evaluated clips are sent to the commercial models for inference only, and TED clips (CC BY-NC-ND) are shown only to the raters and evaluated systems, without clips or records being redistributed.

\subsection{Human Evaluation (Secs.~4.2 and~4.3)}\label{sec:supp_human}

\paragraph{Protocol.}
Three raters who were not involved in this work answered alone in their native language, without seeing any method's output, measurement, or reference, on looping videos that could be paused and replayed; Tab.~\ref{tab:supp_humansets} lists the three item sets. All raters consented to take part and to the release of their anonymized answers.
Besides choosing a video, a rater could answer that they cannot tell, that the question does not fit the videos, or that the body is not visible, and add a note.
Items appear in a fixed random order, the reference video or window comes first in half of the items of Sets 1 and 3, and no feedback is given.
In scoring, the three non-choice answers abstain, an item is labeled with the video chosen by the majority of the non-abstaining raters, and ties or a majority of abstentions leave the item unjudged, so that every label needs at least two agreeing raters.
Sets 1 and 2 name the quantity in plain words with the averaging of both sides, for example ``Who moves the hands faster? (average of the left and right hand; speed of the hands relative to the body, not walking speed)''. Their instructions add that a large movement of one side alone counts for less and that each person is judged relative to their own body. Set 3 poses each pair of windows as an interval comparison of Table~2, with both sides made explicit, for example ``Of the two windows A and B of the same person, in which does the person move both hands faster? (relative to their own body size)'', and repeats four pairs with the sides swapped.
In Set 1 the methods are scored with their answers of Table~1; in Set 2 every setting answers each pair with an English question that adds the plain-language definitions given to the raters; Set 3 uses the answers of Table~2.

\paragraph{Results.}
Tab.~\ref{tab:supp_human} gives the agreement of the methods with the human majority.
In Set 1, 53 items are judged, the unjudged ones being mostly foot-speed questions in which the raters saw no visible movement; the reference agrees with the human majority on 50 of the 53 (94\%, 95\% bootstrap interval [87, 100]), namely 14 of 14 for hand height, 14 of 14 for knee flexion, 10 of 10 for foot speed, and 12 of 15 for hand speed. On these 53 items the Agents agree with the reference on 68--94\% and QMM on 98\%.
Scored against the same human majority, the code that Qwen3.8-27B, Qwen3.5-9B, and Sol write without the definitions agrees on 83, 87, and 94\%, below their frame Agents (96, 91, and 96\%) and QMM (96\%), so programs without shared definitions also agree less with people.
In Set 2, 69 pairs are judged, and the 11 unjudged pairs, six for hand speed and five for hand height, have no majority answer or a majority that could not tell.
Fleiss' kappa including abstentions is 0.91 in Set 1 and 0.58 in Set 2, and raters who both answered agree on 99\% and 86\% of the items.
Paired 95\% bootstrap intervals of the QMM $-$ Agent difference against the human majority lie above zero for Gemma-4-E4B in Set 1 and for Gemma-4-31B, Qwen3.5-9B, Qwen3.5-4B, and Gemma-4-E4B in Set 2, and include zero for all other settings in Set 1 and for Qwen3.8-27B and the commercial settings in Set 2, whose Agents agree with the human majority on 87--88\% of the TED pairs. Measurement-only answers (Sec.~\ref{sec:supp_settings}) agree with the human majority on 96\% (51 of 53) of the mixed-source items and 91\% (63 of 69) of the TED pairs.
In Set 3, all 40 pairs are judged, and the human majority selects the reference window on all 40, namely 20 for the Sol Agent and 20 for the Qwen3.8-27B Agent; QMM with the same backbone answers the reference window on 39 and agrees with the human majority on 39. Fleiss' kappa is 0.81 and 1.00 for the two Agents' pairs, and every rater answered the four swapped pairs consistently.
Of the 16 close comparisons, the raters judge 9 and answer that they cannot tell for most of the others; on the 9, the human majority agrees with the reference and with QMM (Sol at high effort) on all 9 and with the Sol Agent on 4, and on the 5 judged items where the two disagree it sides with QMM each time.
Of the 10 random wrong windows, 8 are judged; the human majority selects the reference window on 7 of them, and QMM answers the reference window on all 8.

\section{Measurement Definitions and Reference Answers (Secs.~3.1, 3.2, and~4.1)}\label{sec:supp_refs}

The definitions below are shared by QMM and by the reference answers, but the reference answers are computed from the 3D motion provided with each dataset and never from the RGB-based records.

\paragraph{Measurement conventions.}
The body scale is the median 3D shoulder width of the video, computed once and used for every interval.
Trajectories are filtered over the whole video before an interval is selected.
Wrist and shoulder trajectories for the hand quantities use a Savitzky--Golay filter (0.5-s window, at least 7 frames, order 5), and the other joints use a fourth-order zero-lag Butterworth filter with cutoff $\min(6\,\mathrm{Hz}, 0.4\,f_s)$ for frame rate $f_s$.
Speeds are central differences on the timestamps.
A measurement is reported only when the interval contains at least 1\,s of valid frames, checked per side, and quantities defined per hand or per leg average the sides that pass.
Peak quantities take the per-frame maximum of each side's filtered signal within the interval and then average the two sides.
Table~\ref{tab:supp_quantities} gives the five quantities, and window search uses a 0.5-s grid of window starts (Sec.~\ref{sec:supp_qmm}).

\begin{table}[t]
\centering
\footnotesize
\setlength{\tabcolsep}{2pt}
\caption{\textbf{Measured quantities.} $\mathbf{w}$, $\mathbf{s}$, $\mathbf{h}$, $\mathbf{k}$, $\mathbf{a}$ are wrist, shoulder, hip, knee, and ankle positions, $\bar{\mathbf{s}}$ and $\bar{\mathbf{h}}$ the shoulder and hip midpoints, $W$ the median shoulder width, $z$ the vertical component, and $\mathrm{mean}_I$ and $\max_I$ are taken over the frames of interval $I$. Side-wise values are averaged over left and right.}
\label{tab:supp_quantities}
\begin{tabular*}{\columnwidth}{@{\extracolsep{\fill}}l l l@{}}
\toprule
Quantity & Definition & Unit \\
\midrule
Hand speed & $\mathrm{mean}_I \lVert \tfrac{d}{dt}(\mathbf{w}-\bar{\mathbf{s}}) \rVert / W$ & $W$/s \\
Foot speed & $\mathrm{mean}_I \lVert \tfrac{d}{dt}(\mathbf{a}-\bar{\mathbf{h}}) \rVert / W$ & $W$/s \\
Hand height & $\max_I (w_z-\bar{s}_z) / W$ & $W$ \\
Stance width & $\mathrm{mean}_I \lVert \mathbf{a}_L-\mathbf{a}_R \rVert / W$ & $W$ \\
Knee flexion & $\max_I \big(180^\circ - \angle(\mathbf{h},\mathbf{k},\mathbf{a})\big)$ & deg \\
\bottomrule
\end{tabular*}
\end{table}

\paragraph{References and near-tie margin.}
Reference values apply the same code to dataset 3D joints, mapped to the joints used by the records (Tab.~\ref{tab:supp_joints}).
The margin $\delta_k = 0.25\,\mathrm{IQR}_k$, a quarter of the interquartile range of quantity $k$, is computed from the HuMMan references of all 616 clips of at least 2\,s and used for every dataset (Tab.~\ref{tab:supp_delta}).
C requires the two values to differ by more than $\delta$, T requires the top value to exceed the second by more than $\delta$, and S leaves candidates within $\delta$ of the anchor unscored.
Int.\ requires the two intervals to differ by more than $\delta$, and Win.\ requires every window starting more than 5\,s from the best one to be at least $\delta$ below it. The conditions for C, T, Int., and Win.\ hold for all but 45 items, which are kept and analyzed below.

\begin{table}[t]
\centering
\footnotesize
\setlength{\tabcolsep}{2pt}
\caption{\textbf{Reference joints.} Shoulder/hip is the ratio of median shoulder width to median hip width in each skeleton, against 2.6--2.9 for the records.}
\label{tab:supp_joints}
\begin{tabular*}{\columnwidth}{@{\extracolsep{\fill}}l l r@{}}
\toprule
Dataset & Reference skeleton & Shoulder/hip \\
\midrule
HuMMan & SMPL~\cite{smpl2015} fits, kinematic joints & 2.45 \\
MoVi & optical motion capture, joint centers & 2.10 \\
Fit3D & Human3.6M~\cite{h36m2014} 17 joints & 1.04 \\
\bottomrule
\end{tabular*}
\end{table}

\begin{table}[t]
\centering
\footnotesize
\caption{\textbf{Near-tie margin} $\delta$ per quantity.}
\label{tab:supp_delta}
\begin{tabular*}{\columnwidth}{@{\extracolsep{\fill}}l r l@{}}
\toprule
Quantity & $\delta$ & Unit \\
\midrule
Hand speed & 0.31 & $W$/s \\
Foot speed & 0.23 & $W$/s \\
Hand height & 0.21 & $W$ \\
Stance width & 0.13 & $W$ \\
Knee flexion & 23.2 & deg \\
\bottomrule
\end{tabular*}
\end{table}

\paragraph{Robustness of the references.}
Tab.~\ref{tab:supp_robust} recomputes every reference answer of Table~1 with one definition changed and reports how often the answer remains unchanged.
Filters change at most 1\% of the answers and the center joint for hand quantities at most 8\%, and reading a quantity as the larger of the two sides instead of their mean keeps 98\% of the answers.

\begin{table}[t]
\centering
\scriptsize
\setlength{\tabcolsep}{1.5pt}
\caption{\textbf{Reference answers under alternative definitions}, as the percentage of answers unchanged (Jaccard for S), pooled over C, T, and S of Table~1, with the values used in the paper in parentheses.}
\label{tab:supp_robust}
\begin{tabular*}{\columnwidth}{@{\extracolsep{\fill}}l rrrr r@{}}
\toprule
Changed definition & HuMMan & MoVi & Fit3D & Mixed & All \\
\midrule
Butterworth 4 / 10\,Hz (6) & 100 / 100 & 100 / 100 & 100 / 100 & 99 / 100 & 100 \\
Sav.--Golay 0.3 / 1.0\,s (0.5) & 100 / 100 & 100 / 99 & 100 / 100 & 100 / 99 & 99--100 \\
Hands rel.\ to hip midpoint & 99 & 99 & 97 & 92 & 97 \\
Larger side, not mean of both & 100 & 93 & 100 & 98 & 98 \\
\bottomrule
\end{tabular*}
\end{table}

\paragraph{Window references under alternative definitions.}
Tab.~\ref{tab:supp_winrobust} recomputes the reference window of every window search with one definition changed, on the same 0.5-s grid; body scales multiply all windows of a video by one constant and cannot move the window.
Stable is the share of searches whose reference window remains within the $\pm$2.5\,s tolerance, and the Agent and QMM columns rescore the stored answers of all eight settings of Table~2 and the seven of Tab.~\ref{tab:supp_fit3d_window} against the changed references.

\begin{table}[t]
\centering
\scriptsize
\setlength{\tabcolsep}{1.5pt}
\caption{\textbf{Window-search references under alternative definitions.} Share of reference windows that remain within the $\pm$2.5\,s tolerance (Stable) and the range over settings of Agent and QMM accuracy against the changed references (120 MoVi and 85 Fit3D searches).}
\label{tab:supp_winrobust}
\begin{tabular*}{\columnwidth}{@{\extracolsep{\fill}}l ccc ccc@{}}
\toprule
& \multicolumn{3}{c}{MoVi} & \multicolumn{3}{c}{Fit3D} \\
\cmidrule(lr){2-4}\cmidrule(lr){5-7}
Changed definition & Stable & Agent & QMM & Stable & Agent & QMM \\
\midrule
Paper definitions & 100 & 2--40 & 95--98 & 100 & 21--52 & 93 \\
Hands rel.\ to hip midpoint & 100 & 2--40 & 95--98 & 93 & 21--55 & 87 \\
Butterworth 4\,Hz (6) & 100 & 2--40 & 95--98 & 100 & 21--52 & 93 \\
Butterworth 10\,Hz (6) & 100 & 2--40 & 95--98 & 100 & 21--52 & 93 \\
Sav.--Golay 0.3\,s (0.5) & 100 & 2--39 & 95--98 & 100 & 21--53 & 93 \\
Sav.--Golay 1.0\,s (0.5) & 98 & 2--38 & 95--97 & 100 & 21--52 & 93 \\
Larger side, not mean of both & 92 & 2--38 & 88--91 & 94 & 21--52 & 87 \\
\bottomrule
\end{tabular*}
\end{table}

\paragraph{Items within the margin.}
In 45 questions of Tables~1 and~2 that involve MoVi foot speed or knee flexion (28 MoVi and 8 mixed-source items of Table~1 and 9 of Table~2), the compared reference values differ by less than the margin; these items are kept, scoring only the other items changes no conclusion, and every paired interval above zero in the main results remains above zero when these items are removed.

\paragraph{Stance width in mixed-source collections.}
Stance width is normalized by shoulder width, and the datasets place the reference shoulder joints differently (Tab.~\ref{tab:supp_joints}), so cross-video questions on mixed-source collections do not include stance width.
Mixed-source chains keep their stance-width step, which compares two intervals of the same video.

\section{QMM Implementation (Sec.~3)}\label{sec:supp_qmm}

\paragraph{Motion records.}
Records are estimated with GVHMR~\cite{gvhmr2024} from each monocular RGB video.
A record holds the joints of the SMPL-X body model~\cite{smplx2019} in world and camera coordinates for every frame of the video, and measurements use the gravity-aligned world-frame joints, at the frame rate of the provided video (30\,fps for HuMMan and MoVi, 50\,fps for Fit3D, and 8\,fps for the VidDiff clips).
The shoulders, elbows, wrists, hips, knees, and ankles are used.
Timestamps are taken from the video, and a joint is valid in a frame when its estimate is finite.
For mixed-source collections, each 2-s or 6-s clip is cut from the record of its source video, and normalization and filtering are recomputed over the clip.

\paragraph{Operators and interface.}
Table~\ref{tab:supp_tools} lists the tools.
Window search evaluates windows of the requested duration whose start lies on a 0.5-s grid from the beginning of the video and whose end lies within it, skips windows whose measurement is unavailable, and returns the earliest window with the highest value.

\begin{table}[t]
\centering
\scriptsize
\setlength{\tabcolsep}{2pt}
\caption{\textbf{Tools.} Both methods have the shared tools; +~QMM additionally has the QMM tools.}
\label{tab:supp_tools}
\begin{tabular*}{\columnwidth}{@{\extracolsep{\fill}}>{\raggedright\arraybackslash}p{0.5\columnwidth} >{\raggedright\arraybackslash}p{0.46\columnwidth}@{}}
\toprule
Tool (arguments) & Returns \\
\midrule
\multicolumn{2}{@{}l}{\textit{Shared}} \\
\texttt{observe}(video, start, end, frames) & 2--16 uniformly sampled frames, 672\,px wide \\
\texttt{compute}(op, values) & sort, max, min, count above a threshold, or pairwise comparison of given numbers \\
\texttt{note\_write}, \texttt{note\_read} & notes kept across questions on the same collection \\
\texttt{final\_answer}(answer) & submits the answer \\
\midrule
\multicolumn{2}{@{}l}{\textit{QMM}} \\
\texttt{qmm\_measure}(videos, quantity, start, end or action) & value per video over the interval or named action, or ``unavailable'' \\
\texttt{qmm\_peak\_window}(video, quantity, window) & start, end, and value of the highest window \\
\texttt{qmm\_list\_formulations}() & quantity names with one-line meanings \\
\texttt{qmm\_describe}(quantity) & definition, joints, unit, and normalization \\
\bottomrule
\end{tabular*}
\end{table}

\section{Compared Methods and Settings (Sec.~4.1)}\label{sec:supp_settings}

\paragraph{Interaction.}
Each episode allows $24N$ frames for $N$ videos, pooled over the turns of a chain, and $\max(24, N{+}16)$ tool calls per turn, with one call per step.
A final answer is requested when the step limit is reached.
At most 256 frames remain in the context; when a request would exceed the context window, the oldest frames are removed first and then older tool results are shortened, under the same rule for both methods.

\paragraph{Backbones.}
Open models are served with vLLM~0.30.0 at temperature 0, 600 output tokens per step, and a 65k context. The checkpoints are Qwen/Qwen3.8-27B-FP8, google/gemma-4-31B-it-qat-w4a16-ct, Qwen/Qwen3.5-9B, Qwen/Qwen3.5-4B, and google/gemma-4-E4B-it from Hugging Face, with thinking disabled through the chat-template flag; the commercial models are gpt-6.1-sol and gemini-3.8-flash, accessed through their APIs in November 2026.
GPT-6.1 Sol runs through the Responses API at low or high reasoning effort with high image detail and 16k output tokens, and Gemini 3.8 Flash runs through its OpenAI-compatible chat endpoint at low reasoning effort with 4k output tokens.
Each setting is run once, and Tab.~\ref{tab:supp_repeat} repeats three Agent settings and two conditions of Table~3(A); batched serving makes repeated runs differ slightly even at temperature 0.

\paragraph{Further conditions.}
Measurement-only answers apply the quantity, scope, and answer rule intended by each question (the larger value, the highest value, the videos above the anchor video, or the highest window) to the measurements on the records, without a VLM. The full-definition Agent receives in its system prompt the definition that QMM returns for every quantity (Tab.~\ref{tab:supp_fulldef}), and the frame-budget sweep raises the budget of the Sol Agent at low effort to 48, 96, 144, and 192 frames per video on the window searches (Tab.~\ref{tab:supp_budget}).

\paragraph{Design analyses.}
In Table~3(B), the VLM runs Python with a function \texttt{measure}(video, quantity, start, end) identical to the interval operator and no search tool. Every row uses the same backbone, questions, and frame access, and only what \texttt{measure} reads changes. It reads one whole-video value per quantity, the time and value of the highest 1-s window per quantity (returned when the requested interval contains that time), values of consecutive 2.5-s or 1-s bins, or the full record. For bins, mean quantities average the bins that overlap the interval, weighted by frames and overlap, and peak quantities take the maximum over bins at least half inside it.
In Tab.~\ref{tab:supp_rgb}, the no-RGB setting removes the frame tool and keeps the records and QMM tools.
Cond.\ Set uses 8-video pools with three exercises and asks which videos of a named exercise exceed an anchor video, so that selecting a video of another exercise counts against the answer, and Vid+Act asks for the top video and its exercise among five options.
For the text-memory control, the same backbone writes one note per clip from eight uniformly sampled frames, with 0--10 estimates of the five quantities, and answers each question from the notes in one text-only call.
The 60 reworded HuMMan questions (20 per task) replace each quantity phrase by one alternative phrase, and an answer change is a different set of selected videos.
MotionBERT records lift RTMW~\cite{rtmw2024} 2D keypoints with the Human3.6M MotionBERT checkpoint, take the camera's up axis as vertical and joint validity from the 2D keypoint confidence, and are measured with the same operators.

\paragraph{Dense video and joint trajectories.}
The dense-video baseline gives the VLM the whole MoVi video at one frame per second (448\,px, each frame preceded by its timestamp) or as one $2{\times}2$ grid of four 224-px frames per second, in a single context without tools; it runs Qwen3.8-27B on window searches and chains and Gemini 3.8 Flash, at low reasoning effort, on window searches.
The joint-trajectory baseline gives the same motion records in the prompt, with the 3D positions in centimeters of the wrists, shoulders, hips, knees, and ankles at 2\,Hz for MoVi videos, which keeps a whole video at about 20--30k tokens within the 65k context of the open models, and at 10\,Hz for the 2-s clips of mixed-source collections, without frames or tools.
Open models run greedily with thinking disabled, and GPT-6.1 Sol runs at high reasoning effort with up to 32k output tokens.

\paragraph{Thinking-enabled Agents.}
With thinking enabled through the chat-template flag and up to 8k output tokens per step, all else as in the main rows, the Qwen3.8-27B and Qwen3.5-9B Agents reach 18\% and 6\% on the 120 MoVi window searches, against 12\% and 6\% with thinking disabled and 98\% with QMM, and 34\% and 27\% on the 85 Fit3D window searches, against 40\% and 28\% without thinking and 93\% with QMM.

\paragraph{Code over the raw joints.}
In the code conditions of Sec.~4.4, the VLM keeps the frame tool and receives one Python tool, \texttt{run\_program}, whose code reads the raw records. \texttt{times(v)} gives the timestamps, \texttt{joint(v, name)} the unfiltered 3D position of a named joint in meters with NaN where it is invalid, and \texttt{duration(v)} the length of the video.
NumPy, SciPy, and the joint names are available, the program has no measurement or search function, and only its printed output, up to 2{,}000 characters, returns to the VLM, together with the type and message of any error; a program is stopped after 120\,s.
With the definitions, the prompt adds the measurement definitions as text, namely the body scale, the filters, central differences for speeds, the 1-s validity rule, and the formula of each quantity.
Without them, the VLM receives only the question, which states the quantity and its normalization as for the Agent.
The interaction budget is that of the other methods.
The open models generate up to 4{,}000 output tokens per step in these conditions, and Sol keeps its 16k output limit.

\paragraph{Timing.}
Table~\ref{tab:supp_cost} uses 45 mixed-source questions (15 each of C, T, and S) and 20 MoVi window searches.
Each backbone runs on one RTX A6000, one question at a time, with records built and the server restarted before each method.
Latency is the wall-clock time of the episode, and unanswered questions count as wrong and are excluded from latency.

\section{Additional Results (Secs.~4.2--4.5)}\label{sec:supp_results}

The human judgments of Secs.~4.2 and~4.3 are reported in Sec.~\ref{sec:supp_human}.

\subsection{Cross-Video Quantitative Analysis (Sec.~4.2)}\label{sec:supp_res_cross}

\paragraph{Confidence intervals.}
Tables~\ref{tab:supp_ci1} and~\ref{tab:supp_ci2} give the +~QMM $-$ Agent difference with 95\% paired-bootstrap intervals (10{,}000 resamples of candidate pools for Table~1, and of videos or mixed-chain collections for Table~2). The resampled units are the 160 mixed-source pools, 75 HuMMan, 10 MoVi, and 32 Fit3D candidate pools of Table~1 and the 80 MoVi videos and 41 mixed-chain collections of Table~2. The intervals resample questions, not runs; repeated runs are reported in Tab.~\ref{tab:supp_repeat}. Differences are computed on unrounded scores, and an interval counts as above zero only when its rounded lower bound is positive.

\paragraph{Input cost.}
In Fig.~3, QMM uses 7.8 to 40.8 times less input per mixed-source top-1 question than the Agent with the same backbone, from 7.8 times for Gemma-4-E4B to 40.8 times for Sol at high effort.

\begin{table*}[t]
\centering
\scriptsize
\setlength{\tabcolsep}{1.5pt}
\caption{\textbf{+~QMM $-$ Agent on Table~1} with 95\% paired-bootstrap intervals.}
\label{tab:supp_ci1}
\begin{tabular*}{\textwidth}{@{\extracolsep{\fill}}l rrr rrr rrr rrr r@{}}
\toprule
& \multicolumn{3}{c}{Mixed} & \multicolumn{3}{c}{HuMMan} & \multicolumn{3}{c}{MoVi} & \multicolumn{3}{c}{Fit3D} & \\
\cmidrule(lr){2-4}\cmidrule(lr){5-7}\cmidrule(lr){8-10}\cmidrule(lr){11-13}
Model & C & T & S & C & T & S & C & T & S & C & T & S & VidDiff \\
\midrule
Qwen-27B & $+5$ {\tiny[$-2$, 11]} & $+21$ {\tiny[6, 35]} & $+20$ {\tiny[8, 32]} & $+25$ {\tiny[15, 35]} & $+48$ {\tiny[32, 64]} & $+49$ {\tiny[38, 61]} & $+20$ {\tiny[12, 27]} & $+50$ {\tiny[42, 58]} & $+46$ {\tiny[36, 55]} & $+24$ {\tiny[15, 33]} & $+30$ {\tiny[13, 49]} & $+30$ {\tiny[22, 37]} & $+25$ {\tiny[15, 36]} \\
Gemma-31B & $+5$ {\tiny[0, 11]} & $+19$ {\tiny[8, 31]} & $+14$ {\tiny[6, 21]} & $+20$ {\tiny[10, 31]} & $+43$ {\tiny[27, 59]} & $+36$ {\tiny[25, 47]} & $+23$ {\tiny[13, 35]} & $+53$ {\tiny[38, 65]} & $+45$ {\tiny[36, 56]} & $+24$ {\tiny[15, 34]} & $+28$ {\tiny[11, 46]} & $+24$ {\tiny[17, 31]} & $+7$ {\tiny[$-1$, 15]} \\
Qwen-9B & $+11$ {\tiny[3, 19]} & $+27$ {\tiny[12, 42]} & $+19$ {\tiny[10, 28]} & $+47$ {\tiny[29, 63]} & $+55$ {\tiny[41, 68]} & $+74$ {\tiny[64, 83]} & $+35$ {\tiny[18, 52]} & $+55$ {\tiny[40, 70]} & $+69$ {\tiny[61, 77]} & $+35$ {\tiny[22, 49]} & $+18$ {\tiny[0, 36]} & $+55$ {\tiny[47, 64]} & $+34$ {\tiny[24, 44]} \\
Qwen-4B & $+8$ {\tiny[2, 14]} & $+35$ {\tiny[21, 50]} & $+14$ {\tiny[1, 27]} & $+37$ {\tiny[23, 51]} & $+48$ {\tiny[34, 62]} & $+56$ {\tiny[44, 66]} & $+27$ {\tiny[15, 40]} & $+50$ {\tiny[35, 63]} & $+69$ {\tiny[57, 80]} & $+30$ {\tiny[19, 42]} & $+25$ {\tiny[9, 43]} & $+45$ {\tiny[38, 52]} & $+41$ {\tiny[31, 51]} \\
Gemma-E4B & $+30$ {\tiny[19, 41]} & $+35$ {\tiny[21, 50]} & $+28$ {\tiny[15, 42]} & $+50$ {\tiny[39, 62]} & $+60$ {\tiny[45, 74]} & $+45$ {\tiny[33, 56]} & $+53$ {\tiny[40, 67]} & $+77$ {\tiny[72, 82]} & $+58$ {\tiny[44, 71]} & $+36$ {\tiny[27, 46]} & $+42$ {\tiny[23, 63]} & $+35$ {\tiny[23, 45]} & $+48$ {\tiny[35, 60]} \\
Sol (low) & $+5$ {\tiny[$-2$, 11]} & $+15$ {\tiny[4, 27]} & $+4$ {\tiny[$-4$, 11]} & $+12$ {\tiny[4, 21]} & $+25$ {\tiny[12, 38]} & $+27$ {\tiny[17, 37]} & $+12$ {\tiny[5, 18]} & $+37$ {\tiny[25, 48]} & $+24$ {\tiny[16, 33]} & $+15$ {\tiny[6, 23]} & $+15$ {\tiny[5, 27]} & $+18$ {\tiny[12, 24]} & $-1$ {\tiny[$-6$, 3]} \\
Sol (high) & $+5$ {\tiny[$-2$, 11]} & $+12$ {\tiny[2, 25]} & $+4$ {\tiny[$-4$, 11]} & $+10$ {\tiny[3, 19]} & $+13$ {\tiny[2, 25]} & $+20$ {\tiny[9, 32]} & $+12$ {\tiny[3, 20]} & $+32$ {\tiny[20, 45]} & $+23$ {\tiny[15, 31]} & $+9$ {\tiny[2, 18]} & $+8$ {\tiny[$-10$, 28]} & $+13$ {\tiny[6, 20]} & $0$ {\tiny[$-6$, 5]} \\
Gemini Flash & $+12$ {\tiny[5, 22]} & $+21$ {\tiny[8, 33]} & $+18$ {\tiny[7, 29]} & $+23$ {\tiny[16, 30]} & $+23$ {\tiny[8, 39]} & $+34$ {\tiny[23, 43]} & $+20$ {\tiny[12, 30]} & $+43$ {\tiny[28, 58]} & $+43$ {\tiny[33, 52]} & $+29$ {\tiny[19, 40]} & $+28$ {\tiny[10, 46]} & $+38$ {\tiny[30, 46]} & $+3$ {\tiny[$-3$, 10]} \\
\bottomrule
\end{tabular*}
\end{table*}

\begin{table*}[t]
\centering
\scriptsize
\setlength{\tabcolsep}{1.5pt}
\caption{\textbf{+~QMM $-$ Agent on Table~2} with 95\% paired-bootstrap intervals over videos (Int., Win., Chain) or mixed-chain collections (Quant., Act., Full).}
\label{tab:supp_ci2}
\begin{tabular*}{0.7\textwidth}{@{\extracolsep{\fill}}l rrr rrr@{}}
\toprule
Model & Int. & Win. & Chain & Quant. & Act. & Full \\
\midrule
Qwen-27B & $+18$ {\tiny[12, 25]} & $+85$ {\tiny[79, 91]} & $+84$ {\tiny[78, 90]} & $+73$ {\tiny[59, 86]} & $+46$ {\tiny[27, 66]} & $+71$ {\tiny[55, 85]} \\
Gemma-31B & $+19$ {\tiny[12, 26]} & $+88$ {\tiny[82, 94]} & $+92$ {\tiny[86, 97]} & $+73$ {\tiny[60, 86]} & $+34$ {\tiny[16, 51]} & $+61$ {\tiny[44, 77]} \\
Qwen-9B & $+32$ {\tiny[24, 41]} & $+92$ {\tiny[86, 97]} & $+94$ {\tiny[90, 98]} & $+80$ {\tiny[68, 91]} & $+39$ {\tiny[26, 53]} & $+66$ {\tiny[51, 79]} \\
Qwen-4B & $+41$ {\tiny[32, 50]} & $+96$ {\tiny[92, 99]} & $+98$ {\tiny[96, 100]} & $+68$ {\tiny[50, 85]} & $+27$ {\tiny[8, 46]} & $+51$ {\tiny[33, 69]} \\
Gemma-E4B & $+44$ {\tiny[35, 54]} & $+90$ {\tiny[84, 95]} & $+89$ {\tiny[83, 95]} & $+78$ {\tiny[64, 90]} & $+22$ {\tiny[5, 38]} & $+32$ {\tiny[17, 46]} \\
Sol (low) & $+12$ {\tiny[6, 19]} & $+60$ {\tiny[52, 68]} & $+70$ {\tiny[63, 78]} & $+56$ {\tiny[42, 69]} & $+20$ {\tiny[7, 32]} & $+56$ {\tiny[42, 69]} \\
Sol (high) & $+12$ {\tiny[6, 18]} & $+58$ {\tiny[49, 67]} & $+69$ {\tiny[62, 77]} & $+46$ {\tiny[31, 62]} & $+17$ {\tiny[5, 30]} & $+46$ {\tiny[31, 62]} \\
Gemini Flash & $+12$ {\tiny[7, 19]} & $+62$ {\tiny[54, 70]} & $+75$ {\tiny[67, 83]} & $+68$ {\tiny[55, 81]} & $+46$ {\tiny[29, 63]} & $+68$ {\tiny[55, 81]} \\
\bottomrule
\end{tabular*}
\end{table*}

\begin{table}[t]
\centering
\scriptsize
\setlength{\tabcolsep}{2pt}
\caption{\textbf{Repeated runs.} Mixed gives Agent C/T/S in two runs, items with the same score, and the gain of the QMM row of Table~1 over each run. Window search gives the code-with-definitions and +~QMM rows of Qwen-27B in Table~3(A).}
\label{tab:supp_repeat}
\begin{tabular*}{\columnwidth}{@{\extracolsep{\fill}}l cc c cc@{}}
\toprule
Mixed & Run 1 & Run 2 & Same & Gain 1 & Gain 2 \\
\midrule
Qwen-4B & 89/60/75 & 84/54/74 & 135/160 & +8/+35/+14 & +12/+42/+15 \\
Gemma-E4B & 67/60/51 & 62/50/57 & 128/160 & +30/+35/+28 & +34/+46/+22 \\
Gemma-31B & 92/77/77 & 91/77/79 & 149/160 & +5/+19/+14 & +6/+19/+12 \\
\midrule
Win., Qwen-27B & Run 1 & Run 2 & Same & & \\
\midrule
Code, definitions & 91 & 93 & 111/120 & & \\
QMM & 98 & 98 & & & \\
\bottomrule
\end{tabular*}
\end{table}

\paragraph{Close and clear comparisons.}
Tab.~\ref{tab:supp_close} splits the 280 comparisons of Table~1 by the difference of the reference values, below or above twice the margin.

\begin{table}[t]
\centering
\scriptsize
\setlength{\tabcolsep}{2pt}
\caption{\textbf{Close and clear comparisons} of Table~1 (mixed, HuMMan, MoVi, Fit3D), Agent $\to$ QMM accuracy and the paired difference with its 95\% bootstrap interval. Close pairs are those whose reference values differ by less than twice the margin (140 pairs), and clear pairs are the others (140).}
\label{tab:supp_close}
\begin{tabular*}{\columnwidth}{@{\extracolsep{\fill}}l ll@{}}
\toprule
Model & Close & Clear \\
\midrule
Qwen-27B & 69 $\to$ 95, $+26$ {\tiny[18, 34]} & 86 $\to$ 98, $+12$ {\tiny[7, 18]} \\
Gemma-31B & 71 $\to$ 95, $+24$ {\tiny[17, 31]} & 85 $\to$ 98, $+13$ {\tiny[7, 19]} \\
Qwen-9B & 56 $\to$ 95, $+39$ {\tiny[31, 48]} & 73 $\to$ 98, $+25$ {\tiny[18, 32]} \\
Qwen-4B & 61 $\to$ 95, $+34$ {\tiny[26, 41]} & 80 $\to$ 98, $+18$ {\tiny[11, 25]} \\
Gemma-E4B & 49 $\to$ 94, $+46$ {\tiny[37, 54]} & 61 $\to$ 98, $+37$ {\tiny[29, 46]} \\
Sol (low) & 78 $\to$ 95, $+17$ {\tiny[10, 24]} & 93 $\to$ 98, $+5$ {\tiny[1, 9]} \\
Sol (high) & 82 $\to$ 95, $+13$ {\tiny[6, 19]} & 93 $\to$ 98, $+5$ {\tiny[1, 9]} \\
Gemini Flash & 66 $\to$ 95, $+29$ {\tiny[21, 36]} & 82 $\to$ 98, $+16$ {\tiny[10, 22]} \\
\bottomrule
\end{tabular*}
\end{table}

\paragraph{Small open backbone against the commercial model.}
On the window searches and mixed-source chains of Table~2, Qwen3.5-4B and Gemma-4-E4B with QMM exceed the Sol Agent at high effort by 58 and 55 points on window search and by 46 and 37 points on the quantitative steps, with 95\% paired-bootstrap intervals of [49, 67], [46, 64], [31, 62], and [19, 54].

\paragraph{Repeated runs.}
Tab.~\ref{tab:supp_repeat} repeats the Agent rows of three open backbones on mixed-source collections and the Qwen3.8-27B rows of Table~3(A) on window search.
The Agent's scores change by up to 10 points between runs, whereas every gain of QMM remains positive.

\paragraph{Text notes and rewording.}
Tab.~\ref{tab:supp_notes} gives a text-memory control, in which the VLM answers from notes written from frames, and the reworded questions of Sec.~4.2 on HuMMan.

\begin{table}[t]
\centering
\footnotesize
\setlength{\tabcolsep}{2.5pt}
\caption{\textbf{Text notes and rewording} on HuMMan. Notes and QMM give C/T/S on the HuMMan questions of Table~1 (60 per column) when the same backbone answers from notes written from eight frames per clip or with QMM; every QMM $-$ notes difference has a 95\% paired-bootstrap interval above zero. Changed is the number of the 60 reworded questions whose answer changes (Agent / QMM).}
\label{tab:supp_notes}
\begin{tabular*}{\columnwidth}{@{\extracolsep{\fill}}l ccc@{}}
\toprule
Model & Notes & QMM & Changed \\
\midrule
Qwen-27B & 68/38/28 & 100/88/87 & 30 / 2 \\
Gemma-31B & 75/30/42 & 100/88/87 & 13 / 0 \\
Qwen-9B & 68/28/33 & 100/88/87 & 29 / 0 \\
Qwen-4B & 62/42/29 & 100/88/87 & 27 / 0 \\
Gemma-E4B & 57/30/28 & 98/88/70 & 26 / 9 \\
\bottomrule
\end{tabular*}
\end{table}

\paragraph{Collection size.}
Tab.~\ref{tab:supp_size} gives the collection-size results summarized in Sec.~4.2.
Every paired interval of the top-1 and set gains lies above zero except set selection for Gemma-4-E4B at 32 videos ($+8$ [$-9$, $+25$]).
For Qwen3.8-27B with QMM, top-1 accuracy is 95, 90, 80, and 60 for 8, 16, 32, and 64 videos, and in every error the correct video is ranked second or third by the records.

\begin{table}[t]
\centering
\footnotesize
\setlength{\tabcolsep}{2pt}
\caption{\textbf{Collection size on HuMMan}, 20 top-1 (T) and 20 set (S) questions per size. Tok.\ is the mean cumulative input per top-1 question (thousands). Bold marks the better accuracy.}
\label{tab:supp_size}
\begin{tabular*}{\columnwidth}{@{\extracolsep{\fill}}l r rr rr rr@{}}
\toprule
& & \multicolumn{2}{c}{T} & \multicolumn{2}{c}{S} & \multicolumn{2}{c}{Tok.} \\
\cmidrule(lr){3-4}\cmidrule(lr){5-6}\cmidrule(lr){7-8}
Model & $N$ & Agent & QMM & Agent & QMM & Agent & QMM \\
\midrule
Qwen-27B & 8 & 65 & \textbf{95} & 67 & \textbf{96} & 183 & 9.5 \\
 & 16 & 55 & \textbf{90} & 58 & \textbf{83} & 457 & 12.4 \\
 & 32 & 15 & \textbf{80} & 38 & \textbf{87} & 1106 & 17.8 \\
 & 64 & 5 & \textbf{60} & 31 & \textbf{79} & 2993 & 108.2 \\
\midrule
Gemma-31B & 8 & 45 & \textbf{95} & 74 & \textbf{96} & 86 & 4.2 \\
 & 16 & 40 & \textbf{90} & 62 & \textbf{83} & 313 & 4.8 \\
 & 32 & 15 & \textbf{80} & 36 & \textbf{87} & 464 & 6.0 \\
 & 64 & 0 & \textbf{60} & 14 & \textbf{89} & 212 & 152.4 \\
\midrule
Qwen-9B & 8 & 40 & \textbf{95} & 53 & \textbf{96} & 228 & 6.1 \\
 & 16 & 55 & \textbf{90} & 49 & \textbf{78} & 571 & 7.8 \\
 & 32 & 10 & \textbf{80} & 35 & \textbf{82} & 1648 & 27.9 \\
 & 64 & 20 & \textbf{60} & 35 & \textbf{78} & 3344 & 43.3 \\
\midrule
Qwen-4B & 8 & 50 & \textbf{95} & 44 & \textbf{96} & 185 & 6.6 \\
 & 16 & 65 & \textbf{90} & 57 & \textbf{83} & 630 & 6.6 \\
 & 32 & 30 & \textbf{75} & 23 & \textbf{72} & 1537 & 14.7 \\
 & 64 & 5 & \textbf{55} & 38 & \textbf{79} & 3416 & 149.1 \\
\midrule
Gemma-E4B & 8 & 20 & \textbf{95} & 40 & \textbf{82} & 119 & 10.2 \\
 & 16 & 30 & \textbf{90} & 21 & \textbf{60} & 395 & 10.4 \\
 & 32 & 25 & \textbf{80} & 18 & \textbf{26} & 1127 & 20.9 \\
 & 64 & 0 & \textbf{65} & 8 & \textbf{44} & 1807 & 35.9 \\
\bottomrule
\end{tabular*}
\end{table}

\paragraph{VidDiff.}
Tab.~\ref{tab:supp_viddiff} breaks VidDiff down by category.
Of the 88 statements, 19 have record values that differ by less than $\delta$, and they contain 4 of the 5 disagreements of QMM in every setting except Gemma-4-E4B (4 of 7).

\begin{table}[t]
\centering
\footnotesize
\setlength{\tabcolsep}{3pt}
\caption{\textbf{VidDiff by category}, Agent / QMM agreement with the annotations (\%).}
\label{tab:supp_viddiff}
\begin{tabular*}{\columnwidth}{@{\extracolsep{\fill}}l ccc@{}}
\toprule
Model & Stance (69) & Knee (19) & All (88) \\
\midrule
Qwen-27B & 62 / 94 & 95 / 95 & 69 / 94 \\
Gemma-31B & 86 / 94 & 95 / 95 & 88 / 94 \\
Qwen-9B & 54 / 94 & 84 / 95 & 60 / 94 \\
Qwen-4B & 48 / 94 & 74 / 95 & 53 / 94 \\
Gemma-E4B & 41 / 94 & 58 / 84 & 44 / 92 \\
Sol (low) & 96 / 94 & 95 / 95 & 95 / 94 \\
Sol (high) & 94 / 94 & 95 / 95 & 94 / 94 \\
Gemini Flash & 91 / 94 & 89 / 95 & 91 / 94 \\
\bottomrule
\end{tabular*}
\end{table}

\paragraph{Motion source.}
Tab.~\ref{tab:supp_observer} compares measurement-only answers, read directly from the measurements without a VLM, for records from GVHMR and from MotionBERT (Sec.~4.2).
MoVi motion capture is part of the GVHMR training data, but MoVi records match their references in absolute terms about as closely as HuMMan records do (ICC .85--.97 against .83--.97, Tab.~\ref{tab:supp_agree}), and the TED clips of the human evaluation lie outside this training data.
Tab.~\ref{tab:supp_agree} gives the absolute agreement of record and reference values on the candidate clips of Table~1.

\begin{table}[t]
\centering
\footnotesize
\caption{\textbf{Motion source.} C/T/S of measurement-only answers with the operators fixed. Bold marks the better estimator. Mixed-source collections exclude stance width.}
\label{tab:supp_observer}
\begin{tabular*}{\columnwidth}{@{\extracolsep{\fill}}l cc@{}}
\toprule
Data & MotionBERT & GVHMR (default) \\
\midrule
Mixed & 86/81/75 & \textbf{97/96/91} \\
HuMMan & 93/60/72 & \textbf{100/88/87} \\
MoVi & 95/83/95 & \textbf{98/90/97} \\
Fit3D & 88/38/62 & \textbf{93/62/73} \\
\bottomrule
\end{tabular*}
\end{table}

\begin{table}[t]
\centering
\scriptsize
\setlength{\tabcolsep}{2pt}
\caption{\textbf{Agreement of record and reference values} on the candidate clips of Table~1, for MoVi over the annotated action intervals that its questions use. Entries give ICC(2,1) / Spearman / mean absolute error (shoulder widths per second for speeds, shoulder widths for height and stance, degrees for knee flexion).}
\label{tab:supp_agree}
\begin{tabular*}{\columnwidth}{@{\extracolsep{\fill}}l cccc@{}}
\toprule
Quantity & HuMMan & MoVi & Fit3D & Mixed \\
\midrule
Hand speed & .97/.97/0.18 & .92/.98/0.31 & .92/.93/0.32 & .96/.97/0.30 \\
Foot speed & .96/.96/0.14 & .94/.98/0.22 & .97/.97/0.19 & .95/.97/0.21 \\
Hand height & .95/.91/0.16 & .95/.99/0.21 & .97/.97/0.23 & .96/.97/0.20 \\
Stance width & .83/.85/0.16 & .85/.93/0.11 & .56/.74/0.30 & .83/.83/0.17 \\
Knee flexion & .95/.95/11.9 & .97/.97/7.7 & .96/.97/11.0 & .96/.90/8.7 \\
\bottomrule
\end{tabular*}
\end{table}

\subsection{Temporal and Multi-Step Analysis (Sec.~4.3)}

\paragraph{Frame budget and search.}
Tab.~\ref{tab:supp_budget} extends the frame budget of Sol at low effort on the 120 window searches.
A window counts as inspected when at least two inspected frames, reconstructed from the observe calls, lie inside the reference 5-s window.
Sol at high effort with 24 frames inspects it in 43\% of the searches and answers 87\% of those correctly, and the Qwen3.8-27B and Qwen3.5-9B Agents inspect it in 33\% and 52\% and answer 35\% and 11\% of those.
When the window is inspected, speed questions are answered at least as accurately as hand height, stance width, and knee flexion.

\begin{table}[t]
\centering
\footnotesize
\setlength{\tabcolsep}{3pt}
\caption{\textbf{Frame budget and search} for the Sol Agent at low effort on the 120 window searches. Inspected is the share of searches in which the reference window was inspected, and Acc.\ if insp.\ the accuracy on those searches. With QMM, Sol at low effort answers 98\% of these searches from the records with 1.9k input tokens.}
\label{tab:supp_budget}
\begin{tabular*}{\columnwidth}{@{\extracolsep{\fill}}r r rr rr@{}}
\toprule
Budget & Win. & Frames & Input (k) & Inspected & Acc.\ if insp. \\
\midrule
24 & 38 & 24 & 20.4 & 39 & 83 \\
48 & 62 & 47 & 45.6 & 66 & 80 \\
96 & 82 & 76 & 105.5 & 96 & 86 \\
144 & 79 & 78 & 106.4 & 91 & 87 \\
192 & 77 & 76 & 101.4 & 93 & 82 \\
\bottomrule
\end{tabular*}
\end{table}

\paragraph{Dense video and joint trajectories.}
Tab.~\ref{tab:supp_samedata} compares the dense-video and joint-trajectory baselines with the Agent and QMM on the window searches and chains of Table~2.
Given the whole video at one frame per second, Gemini 3.8 Flash at low effort reaches 72\% on the 120 window searches with 165k input tokens per question, against 36\% as the Agent with 24 frames and 98\% with QMM.
On mixed-source collections, Sol at high effort given the joint trajectories reaches 92/90/81, against 92/83/87 from frames, 97/96/91 with QMM, and 97/96/89 for Qwen3.5-4B with QMM.

\begin{table}[t]
\centering
\footnotesize
\setlength{\tabcolsep}{3pt}
\caption{\textbf{Dense video and joint trajectories without computation} on MoVi videos (Win./Chain, 120 window searches and 121 chains; -- not run).}
\label{tab:supp_samedata}
\begin{tabular*}{\columnwidth}{@{\extracolsep{\fill}}ll c@{}}
\toprule
Model & Input & Win. / Chain \\
\midrule
Qwen-27B & Agent, 24 frames & 12 / 14 \\
 & Whole video, 1\,fps & 58 / 32 \\
 & Whole video, 4-fps grids & 51 / 31 \\
 & Joint trajectories in prompt & 12 / 10 \\
 & QMM & 98 / 98 \\
\midrule
Sol (high) & Agent, 24 frames & 40 / 29 \\
 & Joint trajectories in prompt & 55 / 60 \\
 & QMM & 98 / 98 \\
\midrule
Gemini Flash & Agent, 24 frames & 36 / 23 \\
 & Whole video, 1\,fps & 72 / -- \\
 & QMM & 98 / 98 \\
\bottomrule
\end{tabular*}
\end{table}

\paragraph{Full measurement definitions for the Agent.}
Tab.~\ref{tab:supp_fulldef} gives the Agent the full definition of every quantity in its system prompt, the same content that QMM returns when the VLM asks for a definition (Secs.~4.2 and~4.3).
The definitions raise interval comparison for Sol but do not raise window search for any backbone.

\begin{table*}[t]
\centering
\footnotesize
\setlength{\tabcolsep}{4pt}
\caption{\textbf{Agent given the full measurement definitions} (Full def.). The Agent receives, in its system prompt, the name, meaning, 3D formula and normalization of every quantity that QMM's operators implement, with frames as its only input. Mixed denotes the 160 questions of Table~1 (64/48/48), and MoVi the 120 interval comparisons and 120 window searches of Table~2. The last row of each block is the paired difference with its 95\% bootstrap interval over candidate pools (Mixed) or videos (MoVi).}
\label{tab:supp_fulldef}
\begin{tabular*}{\textwidth}{@{\extracolsep{\fill}}ll rrr rr@{}}
\toprule
& & \multicolumn{3}{c}{Mixed} & \multicolumn{2}{c}{MoVi} \\
\cmidrule(lr){3-5}\cmidrule(lr){6-7}
Model & Condition & C & T & S & Int. & Win. \\
\midrule
GPT-6.1 Sol (high) & Agent & 92 & 83 & 87 & 88 & 40 \\
 & Full def. & 94 & 92 & 86 & 96 & 39 \\
 & QMM & 97 & 96 & 91 & 99 & 98 \\
 & Full def.\ $-$ Agent & $+2$ {\tiny[0, 5]} & $+8$ {\tiny[$-$2, 21]} & $-1$ {\tiny[$-$9, 8]} & $+8$ {\tiny[3, 14]} & $-1$ {\tiny[$-$9, 8]} \\
\midrule
Qwen3.8-27B & Agent & 92 & 75 & 71 & 81 & 12 \\
 & Full def. & 95 & 77 & 76 & 81 & 7 \\
 & QMM & 97 & 96 & 91 & 99 & 98 \\
 & Full def.\ $-$ Agent & $+3$ {\tiny[0, 8]} & $+2$ {\tiny[$-$6, 10]} & $+5$ {\tiny[$-$1, 12]} & $0$ {\tiny[$-$7, 7]} & $-6$ {\tiny[$-$13, 1]} \\
\midrule
Qwen3.5-9B & Agent & 86 & 69 & 72 & 67 & 6 \\
 & Full def. & 84 & 58 & 64 & 62 & 3 \\
 & QMM & 97 & 96 & 91 & 99 & 98 \\
 & Full def.\ $-$ Agent & $-2$ {\tiny[$-$8, 5]} & $-10$ {\tiny[$-$23, 2]} & $-8$ {\tiny[$-$19, 4]} & $-5$ {\tiny[$-$18, 7]} & $-2$ {\tiny[$-$8, 3]} \\
\bottomrule
\end{tabular*}
\end{table*}

\paragraph{Window search on Fit3D.}
Tab.~\ref{tab:supp_fit3d_window} applies the window rule of Table~2 to the Fit3D videos of at least 15\,s, which are shorter than the MoVi videos, so that chance is 32\%.
Missing answers are rare (0--3 of 85).
The full-definition Agents of the five open backbones reach 18--28\%, and measurement-only answers reach 93\%.

\begin{table}[t]
\centering
\scriptsize
\setlength{\tabcolsep}{2pt}
\caption{\textbf{Window search on Fit3D}, whose videos are not in the training data of the pose estimator, with 85 searches in 62 videos of at least 15\,s, built with the window rule of Table~2 (5-s window, $\pm$2.5\,s). The QMM $-$ Agent column gives the paired 95\% bootstrap interval over videos.}
\label{tab:supp_fit3d_window}
\begin{tabular*}{\columnwidth}{@{\extracolsep{\fill}}l rr r@{}}
\toprule
Model & Agent & + QMM & QMM $-$ Agent \\
\midrule
Qwen3.8-27B & 40 & \textbf{93} & $+53$ {\tiny[40, 65]} \\
Gemma-4-31B & 24 & \textbf{93} & $+69$ {\tiny[58, 79]} \\
Qwen3.5-9B & 28 & \textbf{93} & $+65$ {\tiny[53, 76]} \\
Qwen3.5-4B & 21 & \textbf{93} & $+72$ {\tiny[61, 82]} \\
Gemma-4-E4B & 39 & \textbf{93} & $+54$ {\tiny[41, 66]} \\
GPT-6.1 Sol (high) & 52 & \textbf{93} & $+41$ {\tiny[30, 52]} \\
Gemini 3.8 Flash & 48 & \textbf{93} & $+45$ {\tiny[32, 57]} \\
\bottomrule
\end{tabular*}
\end{table}

\paragraph{Temporal grounding.}
TimeLens-8B~\cite{timelens2026}, a temporal-grounding model fine-tuned from Qwen3-VL-8B, receives each video at 2\,fps with the grounding prompt of its model card and the quantity phrased as an event, for example ``the person raises both hands highest''; the midpoint $m$ of the predicted segment gives the window start $\min(\max(m-2.5,0),T-5)$, scored with the tolerance of Table~2.
On annotated MoVi actions, it localizes 146 actions, up to two per video, with R1@0.5 of 42\% (R1@0.3 49\%, mean IoU 32\%) for actions of 4.6\,s in videos of 155\,s (medians).
Tab.~\ref{tab:supp_grounding} gives the window searches. TimeLens finds many hand- and foot-speed and hand-height peaks on MoVi, but rarely stance-width or knee-flexion peaks.

\begin{table}[t]
\centering
\scriptsize
\setlength{\tabcolsep}{2pt}
\caption{\textbf{Temporal grounding on window search.} TimeLens-8B, correct windows per quantity (correct / searches), with All, QMM (all settings), and Chance in percent.}
\label{tab:supp_grounding}
\begin{tabular*}{\columnwidth}{@{\extracolsep{\fill}}l ccccc ccc@{}}
\toprule
& Hand sp. & Foot sp. & Hand ht. & Stance & Knee & All & QMM & Chance \\
\midrule
MoVi & 43/55 & 23/30 & 12/16 & 0/16 & 1/3 & 66 & 95--98 & 4 \\
Fit3D & 10/27 & 12/19 & 9/13 & 3/10 & 3/16 & 44 & 93 & 32 \\
\bottomrule
\end{tabular*}
\end{table}

\subsection{What the VLM Needs (Sec.~4.4)}

\paragraph{Components.}
Tab.~\ref{tab:supp_code} gives every metric of the conditions of Table~3(A); Tab.~\ref{tab:supp_ladder_ci} gives the paired intervals of each step, and Tab.~\ref{tab:supp_errors} the source of the wrong answers.
An observer error gives the same answer as the operators on the same record, and a computation error a different one.
With the operators, the remaining errors are observer errors or missing answers, whereas the additional errors of the code conditions are computation errors.

\begin{table*}[t]
\centering
\footnotesize
\setlength{\tabcolsep}{4pt}
\caption{\textbf{Component conditions of Table~3(A)}. Mixed denotes the 160 questions of Table~1 (C/T/S), MoVi the interval comparison and window search of Table~2, and input the mean cumulative input tokens per window search (thousands).}
\label{tab:supp_code}
\begin{tabular*}{\textwidth}{@{\extracolsep{\fill}}ll rrr rr r@{}}
\toprule
& & \multicolumn{3}{c}{Mixed} & \multicolumn{2}{c}{MoVi} & \\
\cmidrule(lr){3-5}\cmidrule(lr){6-7}
Model & Condition & C & T & S & Int. & Win. & Input \\
\midrule
Qwen-27B & Agent (frames) & 92 & 75 & 71 & 81 & 12 & 19.9 \\
 & Code, no definitions & 80 & 81 & 77 & 90 & 62 & 28.4 \\
 & Code, definitions & 91 & 83 & 82 & 97 & 91 & 30.8 \\
 & QMM & 97 & 96 & 91 & 99 & 98 & 11.5 \\
\midrule
Qwen-9B & Agent (frames) & 86 & 69 & 72 & 67 & 6 & 102.9 \\
 & Code, no definitions & 84 & 60 & 72 & 87 & 58 & 59.7 \\
 & Code, definitions & 77 & 79 & 65 & 77 & 67 & 80.8 \\
 & QMM & 97 & 96 & 91 & 99 & 98 & 4.6 \\
\midrule
Sol (high) & Agent (frames) & 92 & 83 & 87 & 88 & 40 & 24.2 \\
 & Code, no definitions & 91 & 90 & 83 & 93 & 81 & 39.1 \\
 & Code, definitions & 97 & 96 & 91 & 99 & 98 & 11.3 \\
 & QMM & 97 & 96 & 91 & 99 & 98 & 2.4 \\
\bottomrule
\end{tabular*}
\end{table*}

\begin{table}[t]
\centering
\scriptsize
\setlength{\tabcolsep}{2pt}
\caption{\textbf{Each step of Table~3(A)} with 95\% paired-bootstrap intervals (candidate pools for C, videos for Win.).}
\label{tab:supp_ladder_ci}
\begin{tabular*}{\columnwidth}{@{\extracolsep{\fill}}ll ll@{}}
\toprule
Model & Step & C & Win. \\
\midrule
Qwen-27B & + coordinates, code & $-$12 [$-$25, $-$2] & +50 [+40, +60] \\
Qwen-27B & + definitions, code & +11 [+0, +22] & +28 [+18, +38] \\
Qwen-27B & + operators (QMM) & +6 [+2, +12] & +7 [+1, +13] \\
\midrule
Qwen-9B & + coordinates, code & $-$2 [$-$11, +8] & +52 [+41, +62] \\
Qwen-9B & + definitions, code & $-$8 [$-$20, +5] & +9 [$-$2, +19] \\
Qwen-9B & + operators (QMM) & +20 [+11, +30] & +32 [+24, +40] \\
\midrule
Sol (high) & + coordinates, code & $-$2 [$-$8, +5] & +41 [+29, +52] \\
Sol (high) & + definitions, code & +6 [+2, +12] & +18 [+10, +25] \\
Sol (high) & + operators (QMM) & +0 [+0, +0] & +0 [+0, +0] \\
\bottomrule
\end{tabular*}
\end{table}

\begin{table}[t]
\centering
\scriptsize
\setlength{\tabcolsep}{2pt}
\caption{\textbf{Sources of the wrong answers.} Correct / observer / computation / no answer, where an observer error gives the same answer as the operators on the same record and a computation error differs from it (120 window searches, 64 mixed-source comparisons).}
\label{tab:supp_errors}
\begin{tabular*}{\columnwidth}{@{\extracolsep{\fill}}ll cc@{}}
\toprule
Model & Condition & Win. & C \\
\midrule
Qwen-27B & Code, no definitions & 75 / 1 / 43 / 1 & 51 / 2 / 11 / 0 \\
Qwen-27B & Code, definitions & 109 / 2 / 9 / 0 & 58 / 2 / 4 / 0 \\
Qwen-27B & QMM & 117 / 1 / 0 / 2 & 62 / 2 / 0 / 0 \\
\midrule
Qwen-9B & Code, no definitions & 69 / 2 / 49 / 0 & 54 / 2 / 8 / 0 \\
Qwen-9B & Code, definitions & 80 / 1 / 39 / 0 & 49 / 2 / 12 / 1 \\
Qwen-9B & QMM & 118 / 2 / 0 / 0 & 62 / 2 / 0 / 0 \\
\midrule
Sol (high) & Code, no definitions & 97 / 1 / 22 / 0 & 58 / 2 / 4 / 0 \\
Sol (high) & Code, definitions & 118 / 2 / 0 / 0 & 62 / 2 / 0 / 0 \\
Sol (high) & QMM & 118 / 2 / 0 / 0 & 62 / 2 / 0 / 0 \\
\bottomrule
\end{tabular*}
\end{table}

\paragraph{What code without the definitions computes.}
On the 64 mixed-source comparisons, every program whose final answer agrees with its printed output applies one computation to both videos. The computation errors of Qwen3.8-27B, Qwen3.5-9B, and Sol (11, 8, and 4) include hand speed measured in the world frame without removing body motion, knee flexion taken as the interior angle or the vertical knee range, and hand height taken from another reference or axis. Against the human majority on the 53 judged items, QMM minus code without definitions is $+13$ [4, 25], $+9$ [0, 19], and $+2$ [$-4$, 8] points for the three backbones.
Removing one definition from the operators themselves, without a VLM, also lowers accuracy. Knee flexion read as the interior angle answers 25\% of the knee comparisons against 100\% and agrees with the human majority on 4 of 14 judged knee items against 14 of 14, hand speed without removing body motion finds 80\% of the hand-speed windows against 98\%, and hand height without the shoulder reference 69\% of the hand-height windows against 100\%. With these three definitions and the body scale removed together, accuracy falls from 97\% to 72\% on the 64 comparisons and from 98\% to 84\% on window search.

\paragraph{Paraphrased questions.}
Unlike the one-phrase rewording of Tab.~\ref{tab:supp_notes}, Tab.~\ref{tab:supp_freeform} paraphrases 48 questions of Tables~1 and~2 (12 mixed-source comparisons, 12 mixed-source top-1 questions, 16 MoVi window searches, and 8 MoVi interval comparisons, drawn with a fixed seed among items beyond the margin) while keeping their references and answer instructions.
An LLM that is not among the evaluated models rewrote each question from its meaning under a fixed style guide that excludes measurement terms (speed, height, width, flexion, stance) and keeps the video labels, intervals, and durations, for example ``Which 5-second stretch of V01 has the feet moving most briskly relative to the hips, for the person's size?''; every paraphrase preserves the intended quantity.
With QMM, every open backbone requests the intended quantity in all 48 questions.

\begin{table}[t]
\centering
\footnotesize
\setlength{\tabcolsep}{3pt}
\caption{\textbf{Paraphrased questions.} Accuracy on 48 paraphrased questions and on the same questions in template wording (in parentheses), and the share of QMM runs that request the intended quantity.}
\label{tab:supp_freeform}
\begin{tabular*}{\columnwidth}{@{\extracolsep{\fill}}l cc c@{}}
\toprule
Model & Agent & + QMM & Intended quantity \\
\midrule
Qwen-27B & 58 (65) & 98 (98) & 100 \\
Gemma-31B & 62 (67) & 98 (98) & 100 \\
Qwen-9B & 60 (62) & 98 (98) & 100 \\
Qwen-4B & 46 (54) & 98 (98) & 100 \\
Gemma-E4B & 35 (42) & 98 (92) & 100 \\
\bottomrule
\end{tabular*}
\end{table}

\paragraph{Quantities outside the operators.}
The 60 MoVi chains of this test first ask for the 5-s window in which a side-specific wrist or ankle moves fastest relative to the shoulders or hips. They then ask whether a different side-specific quantity, the bend of an elbow or knee or the height of a wrist above the shoulders, is larger in that window or in a fixed disjoint window, first on one side and then on the other. In 24 chains the answer changes between sides, and none of these quantities is among the five defined quantities.
The VLM uses joint-level operations on the same records, namely speed, height relative to a chosen joint, distance, and angle of chosen joints over an interval and a window search over them, either as tools or as Python functions, with the same filters and validity rule.
The correct plan executes the intended operations on the same records, and references use the dataset motion capture (Tab.~\ref{tab:supp_joint}).
With joint-level tools, Gemma-4-E4B often reads the comparison in the wrong direction, which its code avoids.

\begin{table}[t]
\centering
\scriptsize
\setlength{\tabcolsep}{2pt}
\caption{\textbf{Quantities outside the operators} on 60 MoVi chains. Columns give the window found (step 1), the two side-specific comparisons (steps 2 and 3), and all three steps right.}
\label{tab:supp_joint}
\begin{tabular*}{\columnwidth}{@{\extracolsep{\fill}}ll rrrr@{}}
\toprule
Model & Method & Step 1 & Step 2 & Step 3 & All \\
\midrule
Qwen-27B & Agent & 2 & 50 & 52 & 0 \\
& Joint-level tools & 95 & 100 & 98 & 95 \\
& Joint-level code & 92 & 98 & 98 & 92 \\
\midrule
Qwen-4B & Agent & 0 & 47 & 43 & 0 \\
& Joint-level tools & 97 & 100 & 97 & 95 \\
& Joint-level code & 92 & 97 & 95 & 88 \\
\midrule
Gemma-E4B & Agent & 3 & 45 & 57 & 2 \\
& Joint-level tools & 97 & 38 & 38 & 27 \\
& Joint-level code & 87 & 93 & 90 & 77 \\
\midrule
\multicolumn{2}{@{}l}{Correct plan on the records} & 95 & 100 & 98 & 95 \\
\bottomrule
\end{tabular*}
\end{table}

\paragraph{Query-time RGB.}
Tab.~\ref{tab:supp_rgb} removes the frame tool on HuMMan with the records unchanged. The direct quantitative accuracy of QMM is unchanged, whereas tasks that require recognizing an activity lose 9--15 points, so measurements and frames play distinct roles.

\begin{table}[t]
\centering
\footnotesize
\setlength{\tabcolsep}{2.5pt}
\caption{\textbf{Query-time RGB} on HuMMan with the records unchanged. Cond.\ Set restricts set selection to videos of a named exercise and reports Jaccard, and Vid+Act requires both the selected video and its activity to be correct.}
\label{tab:supp_rgb}
\begin{tabular*}{\columnwidth}{@{\extracolsep{\fill}}ll rrrrr@{}}
\toprule
Model & Method & C & T & S & Cond. & Vid+Act \\
\midrule
Qwen-27B & Agent & 75 & 40 & 38 & 57 & 38 \\
& + QMM, no RGB & \textbf{100} & \textbf{88} & \textbf{87} & 73 & 50 \\
& + QMM & \textbf{100} & \textbf{88} & \textbf{87} & \textbf{82} & \textbf{63} \\
\midrule
Qwen-9B & Agent & 53 & 33 & 13 & 34 & 25 \\
& + QMM, no RGB & \textbf{100} & \textbf{88} & \textbf{87} & 64 & 42 \\
& + QMM & \textbf{100} & \textbf{88} & \textbf{87} & \textbf{79} & \textbf{57} \\
\bottomrule
\end{tabular*}
\end{table}

\subsection{Query-Time Cost and Record Reuse (Sec.~4.5)}

\paragraph{Query-time cost and record construction.}
Tab.~\ref{tab:supp_cost} gives the query-time cost for all five open backbones.
Over all 120 window searches of Table~2, Qwen3.8-27B with QMM uses 11.5k input tokens against 19.9k for its Agent (Tab.~\ref{tab:supp_code}).
GVHMR takes 2.2\,s per second of video for two MoVi videos of 155 and 185\,s, a median of 4.0\,s per second for 78 HuMMan clips of 2--14\,s, and 4.2\,s per second for 69 Fit3D videos of 13--41\,s at 50\,fps, so a fixed start-up time dominates short clips; a record stores the body-model joints in world and camera coordinates, 0.08\,MB per second at 30\,fps and 0.14\,MB at 50\,fps.
With these per-question savings, records built for the clips or videos themselves recover their construction time after two or three questions on a collection of six 2-s clips and after 14--31 window searches on a two-minute video, while the accuracy gain holds from the first question.

\begin{table}[t]
\centering
\footnotesize
\setlength{\tabcolsep}{2.5pt}
\caption{\textbf{Query-time cost for the five open backbones} with built records on one RTX A6000, one question at a time. Lat.\ is the median latency in seconds per question, including tool execution, frame decoding, and all VLM calls; Fr.\ and Tok.\ are the mean numbers of inspected frames and input tokens (thousands) per question. Bold marks the lower cost.}
\label{tab:supp_cost}
\begin{tabular*}{\columnwidth}{@{\extracolsep{\fill}}ll rrr rrr@{}}
\toprule
& & \multicolumn{3}{c}{Mixed} & \multicolumn{3}{c}{MoVi windows} \\
\cmidrule(lr){3-5}\cmidrule(lr){6-8}
Model & Method & Lat. & Fr. & Tok. & Lat. & Fr. & Tok. \\
\midrule
Qwen-27B & Agent & 33.0 & 42.0 & 63.0 & 24.7 & 20.4 & \textbf{20.4} \\
& + QMM & \textbf{5.6} & \textbf{2.9} & \textbf{15.6} & \textbf{5.5} & \textbf{3.1} & 23.1 \\
\midrule
Gemma-31B & Agent & 37.4 & 53.3 & 54.1 & 11.7 & 24.0 & 13.0 \\
& + QMM & \textbf{5.6} & \textbf{0.0} & \textbf{4.5} & \textbf{3.0} & \textbf{0.0} & \textbf{2.3} \\
\midrule
Qwen-9B & Agent & 44.5 & 74.7 & 91.6 & 28.1 & 16.0 & 71.9 \\
& + QMM & \textbf{13.5} & \textbf{0.0} & \textbf{6.5} & \textbf{9.0} & \textbf{0.0} & \textbf{4.5} \\
\midrule
Qwen-4B & Agent & 35.0 & 74.7 & 99.7 & 22.0 & 16.0 & 69.5 \\
& + QMM & \textbf{7.9} & \textbf{0.0} & \textbf{5.9} & \textbf{7.4} & \textbf{4.0} & \textbf{18.3} \\
\midrule
Gemma-E4B & Agent & 24.3 & 37.5 & 54.3 & 19.8 & 21.4 & 20.4 \\
& + QMM & \textbf{3.0} & \textbf{0.2} & \textbf{9.0} & \textbf{7.3} & \textbf{0.8} & \textbf{9.5} \\
\bottomrule
\end{tabular*}
\end{table}

\section{Worked Example}\label{sec:supp_example}

Tab.~\ref{tab:supp_trace} follows the example chain of Fig.~2 for GPT-6.1 Sol at low effort.
The six 6-s clips come from MoVi (V01, V03), Fit3D (V02, V05), and HuMMan (V04, V06).
The questions are (1) in which single video the person moves the wrists the fastest, (2) during which 2-s window of that video the hands are highest and which activity is performed, and (3) whether the feet are wider apart in that window (A) or at 3.5--5.5\,s (B).
The correct answers are V01, the window starting at 1.5\,s with throw/catch, and A.
The Agent views each video once at 12 frames and answers the second and third questions without further observation, selecting the window starting at 0\,s and answering B.
QMM measures hand speed for all six videos, retrieves the window, views eight frames of V01 to name the activity, and measures stance width in both intervals.
The Agent uses 72 frames and 174.8k input tokens over 77\,s, and QMM 8 frames and 25.5k tokens over 18\,s.
Of the eight settings, QMM answers the chain correctly in seven and the Agent in three, and Gemma-4-E4B with QMM names the wrong activity.

\begin{table}[t]
\centering
\footnotesize
\setlength{\tabcolsep}{2pt}
\caption{\textbf{QMM calls for the example chain of Fig.~2} (GPT-6.1 Sol, low effort). Values are in shoulder widths per second (step 1) and shoulder widths (steps 2--3), with dataset-3D references in parentheses.}
\label{tab:supp_trace}
\begin{tabular*}{\columnwidth}{@{\extracolsep{\fill}}l >{\raggedright\arraybackslash}p{0.7\columnwidth}@{}}
\toprule
Step & Call and returned value \\
\midrule
1 & \texttt{qmm\_describe}(hand speed) \\
 & $r_1$ \texttt{qmm\_measure}(V01--V06, hand speed, 0--6\,s): V01 2.85 (2.16), V02 0.82 (0.24), V03 1.14 (1.12), V04 0.32 (0.28), V05 1.28 (1.84), V06 0.36 (0.37) $\rightarrow$ V01 \\
\addlinespace
2 & $r_2$ \texttt{qmm\_peak\_window}(V01, hand height, 2\,s): 1.5--3.5\,s, 1.13 (0.93; next best window 0.87 (0.64)) \\
 & \texttt{observe}(V01, 0--6\,s, 8 frames) $\rightarrow$ throw/catch \\
\addlinespace
3 & $r_3$ \texttt{qmm\_measure}(V01, stance width, 1.5--3.5\,s): 1.51 (1.25) \\
 & $r_4$ \texttt{qmm\_measure}(V01, stance width, 3.5--5.5\,s): 1.33 (1.10) $\rightarrow$ A \\
\bottomrule
\end{tabular*}
\end{table}

\section{Prompts}\label{sec:supp_prompts}

\paragraph{System prompt, both methods.}
\begin{quote}\small
You analyse a collection of \{n\} videos (V01--V\{n\}), each \{description\} (\{V01: d s, \ldots\}). You cannot see the videos unless you call observe. Act step by step, one tool call per message; when you know the answer, call final\_answer (or output one JSON object \{\texttt{"final"}: \ldots\}). Quantities are judged relative to each person's own body size. Base every answer on what you observe or measure in the time intervals the question refers to; compare videos only on the requested quantity.
\end{quote}
The description is ``a short clip of one person performing one exercise'' for HuMMan, ``one person performing several repetitions of one gym exercise'' for Fit3D, and ``a video of one person doing a series of actions'' otherwise, including mixed-source collections.
On MoVi questions with action intervals, both methods also receive the annotated action-time table.

\paragraph{Addition for +~QMM.}
\begin{quote}\small
QMM motion memory tools give measured values from the videos' pose records (larger = more; `unavailable' if the measurement is not supported). Available quantities: [quantity names]. Use qmm\_describe for definitions.
\end{quote}
For MoVi questions with action intervals, +~QMM is also told that \texttt{qmm\_measure} accepts an action name and then measures each video within its own span of that action.

Each question ends with a line giving the required answer form, such as a video label, a list of labels, a start time in seconds, or A or B.

\paragraph{Joint-trajectory baseline.}
\begin{quote}\small
System: You analyse the motion of one person from a 3D joint-position record. Answer the question from the record.\\
User: Motion record of the person in V01 estimated from the video (3D joint positions in cm; z is up). One row per 0.5\,s: time in seconds, then x y z for l\_wrist, r\_wrist, l\_shoulder, r\_shoulder, l\_hip, r\_hip, l\_knee, r\_knee, l\_ankle, r\_ankle. `na' = not visible. [rows] [question] [answer format]
\end{quote}
Mixed-source collections give one block per video at 0.1\,s per row.

\paragraph{Dense-video baseline.}
\begin{quote}\small
System: You analyse a video of one person. Answer the question from the frames.\\
User: Video V01, sampled at one frame per second ($n$ frames, 0 to $T$\,s). Each frame is preceded by its timestamp. [t = 0 s, frame, \ldots] [question] [answer format]
\end{quote}

\paragraph{VidDiff questions.}
\begin{quote}\small
Both videos show the exercise `\{action\}' (\{action description\}). In which video is the following more true: \{statement\}? Answer V01 or V02.
\end{quote}

\end{document}